%% file: cvpr.tex
\documentclass[final]{cvpr}
\usepackage{times}
\usepackage{epsfig}
\usepackage{graphicx}
\usepackage{amsmath}
\usepackage{amssymb}

\usepackage{multirow}
\usepackage{booktabs}
\usepackage{color}

\usepackage[accsupp]{axessibility}

\newtheorem{proposition}{Proposition}

\newtheorem{condition}{Condition}

\usepackage[pagebackref=true,breaklinks=true,colorlinks,bookmarks=false]{hyperref}

\def\cvprPaperID{1004} 
\def\confYear{CVPR 2022}
\begin{document}

\title{Online Convolutional Re-parameterization}

\author{Mu Hu\textsuperscript{1}\thanks{Work done during an internship at Alibaba Cloud Computing Ltd.} \quad Junyi Feng\textsuperscript{2} \quad Jiashen Hua\textsuperscript{2} \quad Baisheng Lai\textsuperscript{2}  \\ Jianqiang Huang\textsuperscript{2} \quad Xiaojin Gong\textsuperscript{1}\thanks{Corresponding Author.} \quad Xiansheng Hua\textsuperscript{2} \\ 
\textsuperscript{1} Zhejiang University \quad \textsuperscript{2} Alibaba Cloud Computing Ltd. \\
\tt\small muhu@zju.edu.cn [felix.fjy, jiashen.hjs, baisheng.lbs]@alibaba-inc.com \\ \tt\small
jianqiang.hjq@alibaba-inc.com, gongxj@zju.edu.cn, xiansheng.hxs@alibaba-inc.com
}

\maketitle

\pagestyle{empty}
\thispagestyle{empty}

\input{sections/abstract.tex}
\input{sections/intro.tex}
\input{sections/related.tex}

\input{sections/method.tex}
\input{sections/experiment.tex}
\input{sections/conclusion.tex}

{\small
\bibliographystyle{ieee_fullname}
\bibliography{reference}
}

\end{document}


\title{Supplementary Materials: Online Convolutional Re-parameterization}

\maketitle

\input{sections_supp/method.tex}
\input{sections_supp/experiment.tex}

{\small
\bibliographystyle{ieee_fullname}
\bibliography{reference_supp}
}

%% file: sections/abstract.tex
\begin{abstract}
    Structural re-parameterization has drawn increasing attention
    in various computer vision tasks.
    It aims at improving the performance of deep models without introducing 
    any inference-time cost.
    Though efficient during inference, 
    such models rely heavily on the complicated training-time blocks to 
    achieve high accuracy, leading to large extra training cost.
    In this paper, we present online convolutional re-parameterization~(OREPA),
    a two-stage pipeline, aiming to reduce the huge training overhead by 
    squeezing the complex training-time block into a single convolution.
    To achieve this goal, we introduce a linear scaling layer 
    for better optimizing the online blocks.
    Assisted with the reduced training cost, we also explore
    some more effective re-param components.
    Compared with the state-of-the-art re-param models, OREPA is able to 
    save the training-time memory cost by about 70\% and accelerate 
    the training speed by around 2$\times$. 
    Meanwhile, equipped with OREPA, the models outperform previous 
    methods on ImageNet by up to +0.6\%. 
    We also conduct experiments on object detection and semantic segmentation 
    and show consistent improvements on the downstream tasks. Codes are available at  {\color{red}\url{https://github.com/JUGGHM/OREPA_CVPR2022}}.
    
\end{abstract}

%% file: sections/intro.tex
\section{Introduction}
\label{sec:intro}
Convolutional Neural Networks~(CNNs) have seen the success of many computer vision tasks,
including classification~\cite{He16,Krizhevsky12,Simonyan15,Huang17}, object detection~\cite{Ren15,Lin17,Redmon16}, 
segmentation~\cite{Chen17,Zhao17}, \etc.
The trade-off between accuracy and model efficiency has been widely discussed.
In general, a model with higher accuracy usually requires a more complicated block~\cite{Huang17, Hu18, Alexey21},
a wider or deeper structure~\cite{Zagoruyko17, Brock21nfnet}.
However, such models are always too heavy to be deployed, especially in the scenarios where 
the hardware performance is limited, and real-time inference is required.
Taking efficiency into consideration, smaller, compacter, and faster models are preferred. 

In order to obtain a deploy-friendly model and keep a high accuracy, 
structural re-parameterization based methods \cite{Ding19,Ding21dbb,Ding21repvgg,Guo20} are proposed 
for a ``\emph{free}'' performance improvement. In such methods, the models have different structures
during the training phase and the inference phase. 
Specifically, they~\cite{Ding21dbb,Arora18} use complicated training-phase topologies, 
\ie, re-parameterized blocks, to improve the performance.
After training, they squeeze a complicated block into a single linear layer
through equivalent transformation. 
The squeezed models are usually with a neat architecture,
\eg, usually a VGG-like~\cite{Ding21repvgg} or a ResNet-like~\cite{Ding21dbb} structure.
From this perspective, the re-parameterization strategies can improve model performances
without introducing additional inference-time cost.
\begin{figure}[tbp]
	\begin{center}
		\includegraphics[width=1.0\linewidth]{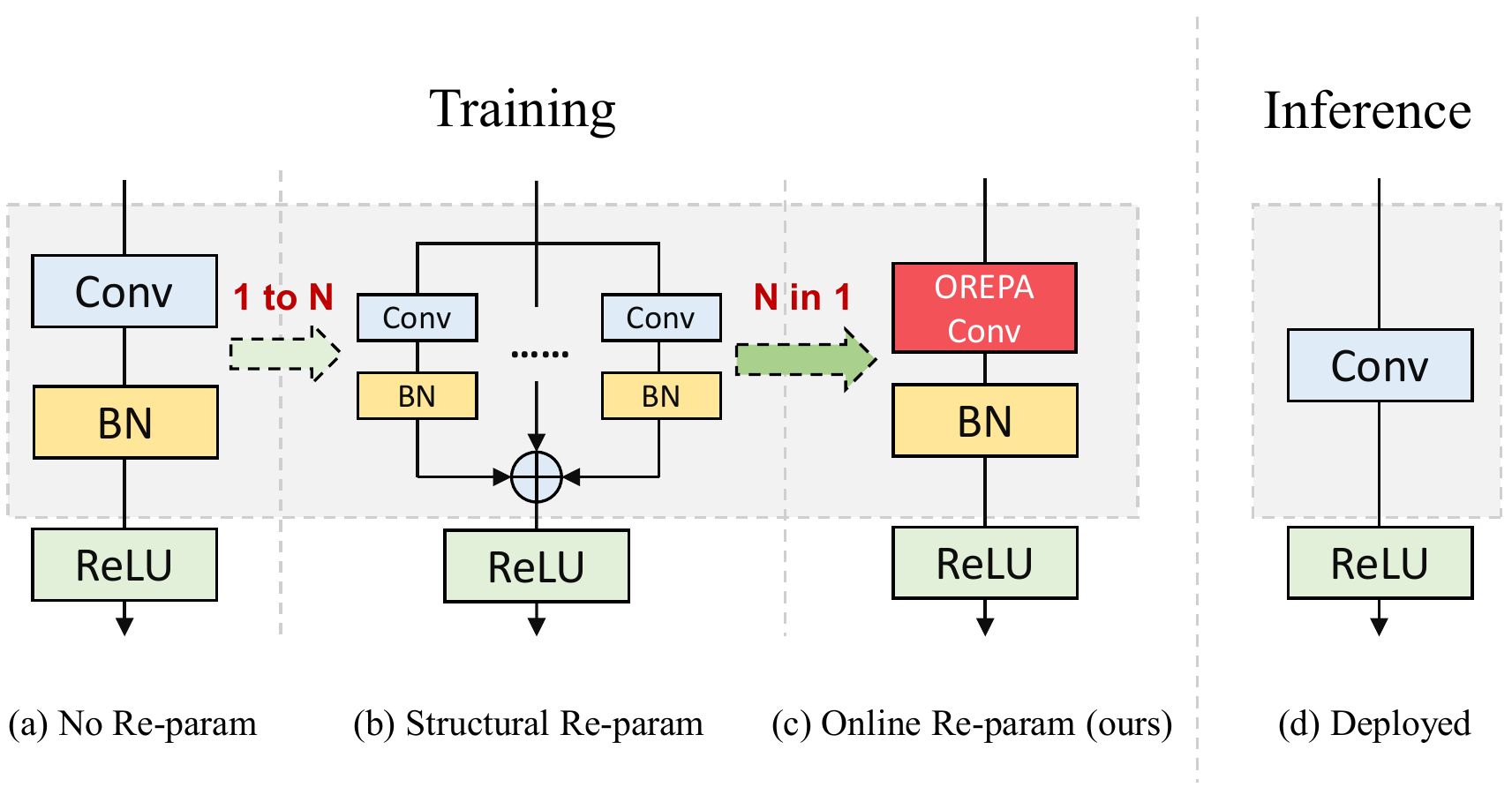}
	\end{center}
	\vspace{-1.em}
	\caption{Comparison of (a) a vanilla convolutional layer, 
	(b) a typical re-param block, and 
	(c) our online re-param block in the training phase. 
	All of these structures are converted to the same (d) inference-time structure.}
	\label{fig:intro}
\end{figure}

It is believed that the normalization~(norm) layer is the crucial component 
in re-param models.
In a re-param block~(\figref{fig:intro}(b)),
a norm layer is always added right-after each computational layer. 
It is observed that the removal of such norm layers would lead to severe performance degradation~\cite{Ding21repvgg, Ding21dbb}.
However, when considering the efficiency, the utilization of such norm layers
unexpectedly brings huge computational overhead in the training phase.
The complicated block could be squeezed into a single convolutional layer in the inference phase.
But, during training, the norm layers are non-linear, \ie, 
they divide the feature map by its standard deviation, 
which prevents us from merging the whole block.
As a result, there exist plenty of intermediate computational operations~(large FLOPS) and buffered feature maps~(high memory usage).
Even worse, the high training budget makes it difficult to explore more complex and potentially stronger re-param blocks.
Naturally, the following question arises,
\begin{itemize}
	\item \textit{Why does normalization matter in re-param?}
\end{itemize}
According to the analysis and experiments, we claim that it is the 
\emph{scaling factors} in the norm layers that counts most,
since they are able to \emph{diversify the optimization direction} of different branches.

Based on the observations, 
we propose Online Re-Parameterization~(OREPA)~(\figref{fig:intro}(c)), a 
two-stage pipeline which enables us to simplify the complicated training-time re-param blocks. 
In the first stage, block linearization, we remove all the non-linear norm layers and introduce 
the \emph{linear scaling layers}. 
Such layers are with similar property as norm layers, that they diversify the optimization of different branches. 
Besides, these layers are linear,
and can be merged into convolutional layers during training. 
The second stage, named block squeezing, 
simplifies the complicated linear block into a single convolutional layer. 
The OREPA significantly shrinks the training cost by reducing the computational and storage overhead caused by 
the intermediate computational layers, with only minor compromising on performance. 
Moreover, the high-efficiency makes it feasible to explore much more complicated re-parameterized topologies. 
To validate this, we further propose several re-parameterized components for better performance. 

We evaluate the proposed OREPA on the ImageNet~\cite{Deng09} classification task.
Compared with the state-of-the-art re-param models~\cite{Ding21dbb}, 
OREPA reduces the extra training-time GPU memory cost by 
65$\%$ to 75$\%$, 
and speeds up the training process by 1.5$\times$ to 2.3$\times$.
Meanwhile, 
our OREPA-ResNet and OREPA-VGG consistently
outperform previous methods~\cite{Ding21dbb, Ding21repvgg}
by +0.2\%$\sim$+0.6\%.
We evaluate OREPA on the downstream tasks, \ie,
object detection and semantic segmentation. 
We find that OREPA could consistently bring performance gain on these tasks.


Our contributions can be summarized as follows:
\begin{itemize}
	\item 
	We propose the Online Convolutional Reparameterization~(OREPA) strategy,
	which greatly improves the training efficiency of re-parameterization models
	and makes it possible to explore stronger re-param blocks.
	
	\item 
	According to our analysis on the mechanism by which the re-param models work,
	we replace the norm layers with the introduced linear scaling layers,
	which still provides diverse optimization directions
	and preserves the representational capacity.

	\item 
	Experiments on various vision tasks 
	demonstrate OREPA outperforms previous re-param models in terms of both
	accuracy and training efficiency.
\end{itemize}

%% file: sections/related.tex
\section{Related Works}
\label{sec:related}

\subsection{Structural Re-parameterization}
Structural re-parameterization \cite{Ding21dbb, Ding21repvgg} 
is recently attached greater importance and utilized in lots of computer vision tasks, 
such as compact model design \cite{Alexey21}, 
architecture search \cite{Chen19, Zhang21}, and pruning \cite{Ding21resrep}. 
Re-parameterization means different architectures can be mutually converted through equivalent transformation of parameters. 
For example, a branch of 1$\times$1 convolution and a branch of 3$\times$3 convolution, 
can be transferred into a single branch of 3$\times$3 convolution \cite{Ding21repvgg}. 
In the training phase,
multi-branch \cite{Ding19, Ding21dbb, Ding21repvgg} and multi-layer \cite{Guo20, Cao20} topologies 
are designed to replace the vanilla linear layers (\eg conv or full connected layer \cite{Arora18}) for augmenting models. 
Cao \etal \cite{Cao20} have discussed how to merge a depthwise separable convolution kernel during training. 
Afterwards during inference, 
the training-time complex models are transferred to simple ones for faster inference. 
While benefiting from complex training-time topologies, 
current re-parameterization methods\cite{Ding19, Ding21dbb, Guo20} 
are trained with non-negligible extra computational cost. 
When the block becomes more complicated for stronger representation, 
the GPU memory utilization and time for training will grow larger and longer, 
finally towards unacceptable.
Different from previous re-param methods, we focus more on the training cost.
We propose a general online convolutional re-parameterization strategy, which make the training-time
structural re-parameterization possible.

\subsection{Normalization}
Normalization \cite{Ioffe15, Wu18, Ba16, Salimans16} is proposed 
to alleviate the gradient vanishing problem when training very deep neural networks. 
It is believed that norm layers are very essential \cite{Santurkar18} as they smooth the loss landscape. 
Recent works on norm-free neural networks claim that 
norm layers are not indispensable \cite{Zhang19, Shao20}. 
Through good initialization and proper regularization, 
normalization layers can be removed \cite{Zhang19, De20, Shao20, Brock21} elegantly. 

\begin{figure*}[tbp]
	\begin{center}
		\includegraphics[width=1.0\linewidth]{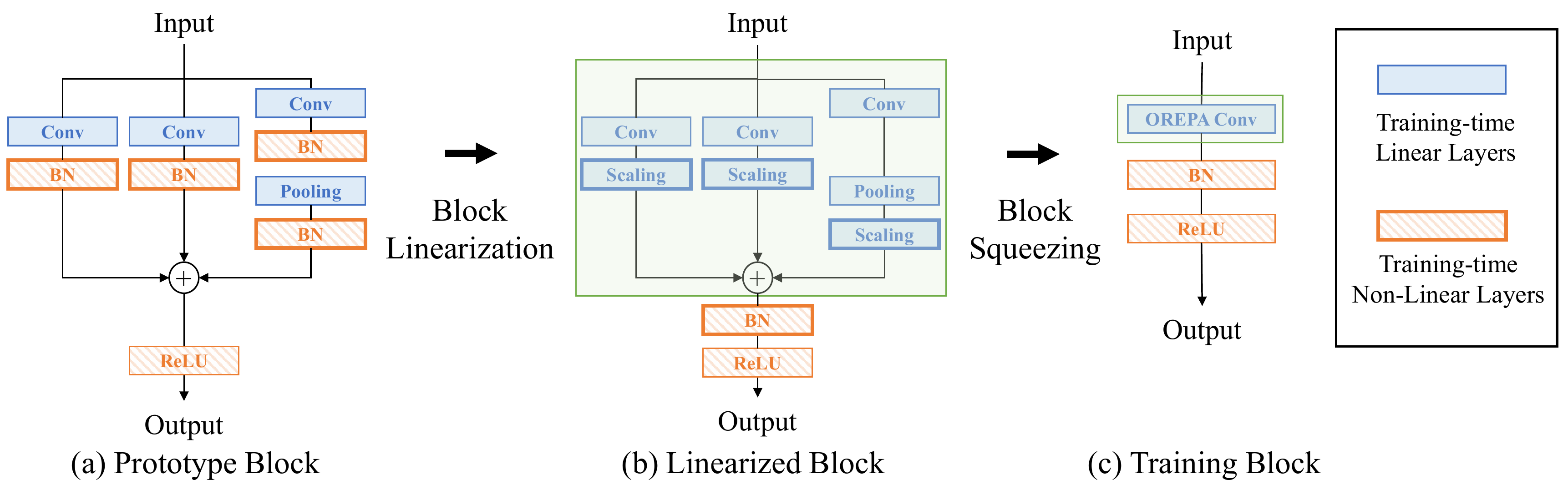}
	\end{center}
	\vspace{-1.em}
	\caption{An overview of the proposed Online Re-Parameterization~(OREPA), a two-stage pipeline.
		In the first stage~(Block Linearization), we remove all the non-linear components in the prototype re-param block.
		In the second stage~(Block Squeezing), we merge the block to a single convolutional layer~(OREPA Conv).
		Through the steps, we significantly reduce the training cost while keep the high performance.
	}
	\label{fig:orepa}
\end{figure*}

For the re-parameterization models, 
it is believed that norm layers in re-parameterized blocks are crucial~\cite{Ding21dbb, Ding21repvgg}. 
The norm-free variants would suffer from performance degradation. 
However, the training-time norm layers are non-linear, \ie, they divide the feature map by its standard deviation, which prevents us from merging the blocks online.
To make it feasible to perform online re-parameterization, we remove all the norm layers in re-param blocks, 
In addition, we introduce the linear alternative of norm layers, \ie, the linear scaling layers.

\subsection{Convolutional Decomposition}
Standard convolutional layers are computational dense, 
leading to large FLOPs and parameter numbers. Therefore, 
convolutional decomposition methods \cite{Mamalet12, Wang17, Romera17} are proposed 
and widely applied in light-weighted models for mobile devices \cite{Howard17, Chollet17}. 
Re-parameterization methods \cite{Ding19, Ding21dbb, Guo20, Cao20, Chen19, Zhang21} 
could be regarded as a certain form of convolutional decomposition as well, 
but towards more complicated topologies. 
The difference of our methods is that we decompose convolution at the kernel level rather than structure level.

%% file: sections/method.tex
\section{Online Re-Parameterization}
\label{sec:methods}
In this section, we introduce the proposed Online Convolutional Re-Parameterization.
First, we analyze the key components, \ie, the normalization layers in re-parameterization models, as preliminaries in \secref{sec:prelim}. 
Based on the analysis, we propose the Online Re-Parameterization~(OREPA), aiming to greatly reduce the training-time budgets of re-parameterization models.
OREPA is able to simplify the complicated training-time block into \emph{\textbf{a single convolutional layer}} and preserve the high accuracy. 
The overall pipeline of OREPA is illustrated in \figref{fig:orepa}, 
which consists of a Block Linearization stage~(\secref{sec:linearization}) 
and a Block Squeezing stage~(\secref{sec:trans}).
Next, in ~\secref{sec:principle}, we dig deeper into the effectiveness of re-parameterization by analyzing the optimization diversity of the multi-layer 
and multi-branch structures, and prove that both the proposed linear scaling layers and normalization layers have the similar effect.
Finally, with the reduced training budget, 
we further explore some more components for a stronger re-parameterization~(\secref{sec:components}), 
with marginally increased cost.

\subsection{Preliminaries: Normalization in Re-param}
\label{sec:prelim}
%
\begin{table}[tbp]
	\caption{
	Effectiveness of normalization layers in re-param models.
	To stabilize the training process, when removing the branch-wise norm layers, 
	we add a post-addtion norm layer.
	}
	\vspace{1.0em}
	\small
	\centering\
	\begin{tabular}{lcc}
		\toprule
		  Variants 								& DBB-18 	& RepVGG-A0 	\\ \midrule				
		  Original								& 71.77	 	& 72.41 		\\
		  W/o branch-wise norm					& 71.35		& 71.15			\\
		  W/o re-param					& 71.21		& 71.17			\\
		  \bottomrule
	\end{tabular}
	\label{tab:norm}
\end{table}
It is believed that the intermediate normalization layers are the key components for 
the multi-layer and multi-branch structures in re-parameterization.
Taking the SoTA models, \ie, DBB~\cite{Ding21dbb} and RepVGG~\cite{Ding21repvgg} as examples,
the removal of such layers would cause severe performance degradation, 
as shown in \tabref{tab:norm}. 
Such an observation is also experimentally supported by Ding \etal\cite{Ding21dbb, Ding21repvgg}.
Thus, we claim that the intermediate normalization layers are \textbf{\emph{essential for performances}} of re-parameterization models.

However, the utilization of intermediate norm layers unexpectedly \textbf{\emph{brings higher training budgets}}.
We notice that in the inference phase, 
all the intermediate operations in the re-parameterization block are linear,
thus can be merged 
into one convolutional layer, resulting in a simple structure.
But during training, the norm layers are non-linear,
\ie, they divide the feature map by its standard deviation. 
As a result, the intermediate operations should be calculated separately, 
which leads to higher computational and memory costs. 
Even worse, such high cost would prevent the community from exploring stronger training blocks.
\begin{figure*}[htbp]
	\begin{center}
		\includegraphics[width=1.0\linewidth]{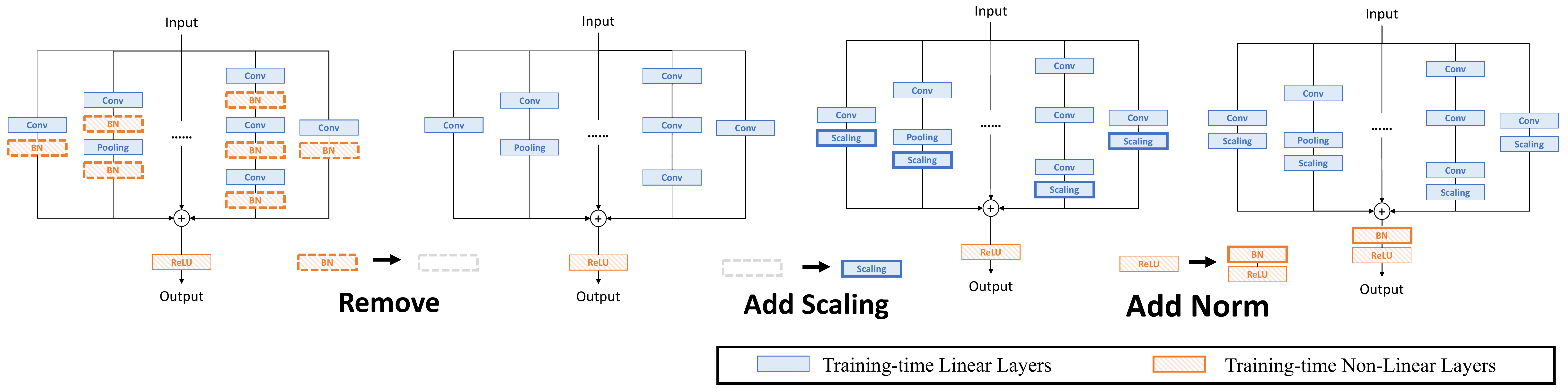}
	\end{center}
	\vspace{-.8em}
	\caption{Three steps of block linearization.
	i) We first \textbf{remove} all the training-time non-linear normalization layers in the ra-param block.
	ii) Second, we add a linear \textbf{scaling} layer at the end of each branch to 
	diversify the optimization directions.
	iii) Last, we add a post-normalization layer right after each block to stabilize training.}
	\label{fig:linearization}
\end{figure*}

\subsection{Block Linearization}
\label{sec:linearization}

As stated in \secref{sec:prelim}, the intermediate normalization layers prevent us from 
merging the separate layers during training. 
However, it is non-trivial to directly remove them due to the performance issue.
To tackle this dilemma, we introduce the channel-wise linear scaling operation
as a linear alternative of normalization.
The scaling layer contains a learnable vector, 
which scales the feature map in the channel dimension.
The linear scaling layers have the similar effect as normalization layers, 
that they both \emph{\textbf{encourage the multi-branches to be optimized towards diverse directions}},
which is the key to the performance improvement in re-parameterization.
The detailed analysis of the effect is discussed in \secref{sec:principle}.
In addition to the effects on performance, the linear scaling layers
could be merged during the training, making the online re-parameterization possible.

Based on the linear scaling layers, we modify the re-parameterization blocks
as illustrated in \figref{fig:linearization}.
Specifically, the block linearization stage consists of the following three steps.
%
First, we remove all the non-linear layers, \ie,
normalization layers in the re-parameterization blocks.
%
Second, in order to maintain the optimization diversity,
we add a scaling layer, the linear alternative of normalization,
at the end of each branch.
%
Finally, to stabilize the training process, 
we add a post-addition normalization layer right after the addition of all the branches.

Once finishing the linearization stage, there exist only linear layers
in the re-param blocks, meaning that
we can merge all the components in the block during the training phase.
Next, we describe how to squeeze such a block into a single convolution kernel. 

\subsection{Block Squeezing}
\label{sec:trans}
Benefiting from block linearization~(\secref{sec:linearization}),
we obtain a linear block.
In this section, we describe the standard procedure for 
squeezing a training-time linear block into a single 
convolution kernel.
The block squeezing step converts the operations on intermediate feature maps, 
which are computation and memory expensive, 
into operations on kernels that are much more economic.
This implies that we reduce the extra training cost of re-param 
from $O(H\times W)$ to $O(K_H\times K_W)$ in terms of 
both computation and memory, 
where $(H,W),\ (K_H, K_W)$ are the spatial shapes of 
the feature map and the convolutional kernel.
%
\begin{figure}[tbp]
	\begin{center}
		\includegraphics[width=1.0\linewidth]{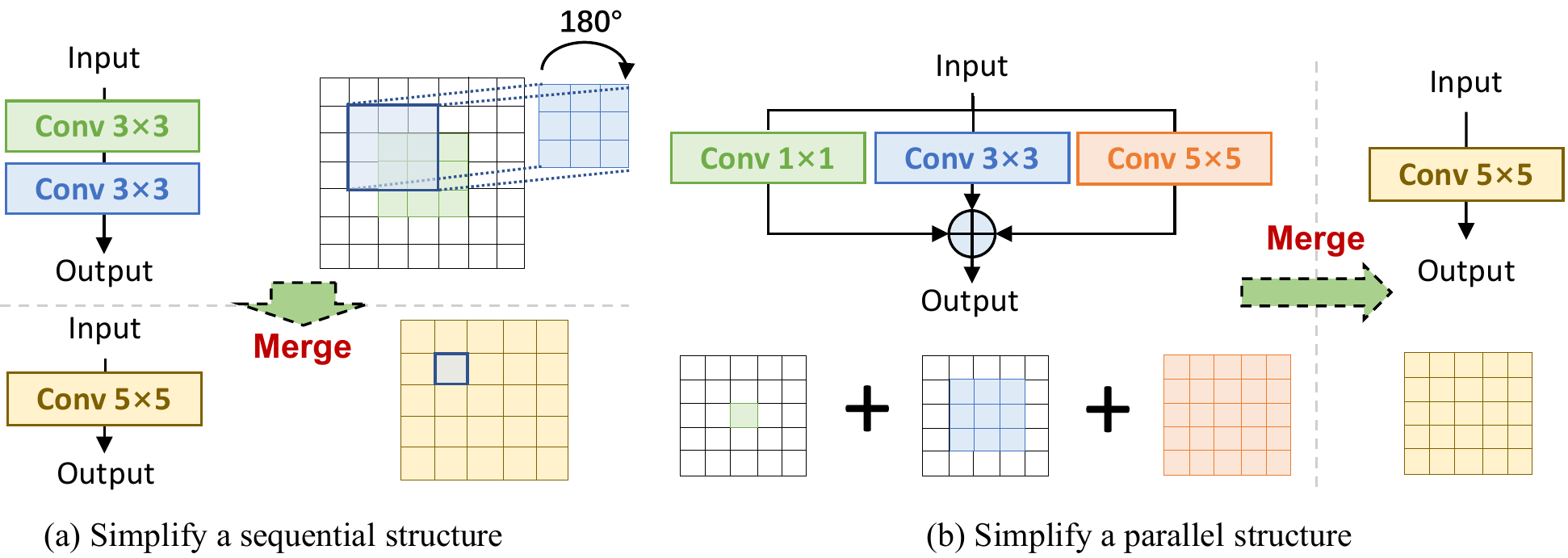}
	\end{center}
	\vspace{-.5em}
	\caption{Simplification of sequential and parallel structures.
	Such simplifications convert the heavy computations on the feature maps to 
	the lighter ones on the convolutional kernels.}
	\label{fig:squeeze}
\end{figure}

In general, no matter how complicated a linear re-param block is, 
the following two properties always hold.
\begin{itemize}
	\item All the linear layer in the block, \eg, 
	depth-wise convolution, average pooling, and the proposed linear scaling,
	can be represented by a degraded convolutional layer with a corresponding set 
	of parameters. Please refer to the supplementary materials for details. 
	\item The block can be represented by a series of parallel branches, 
	each of which consists a sequence of convolutional layers.
\end{itemize}

With the above two properties, we can squeeze a block if we can simplify both
i) a multi-layer~(\ie, the sequential structure) and
ii) a multi-branch~(\ie, the parallel structure) into a single convolution.
In the following part, 
we show how to simplify the sequential structure~(\figref{fig:squeeze}(a))
and the parallel structure~(\figref{fig:squeeze}(b)).

We first define the notations of convolution.
Let $C_{i}$, $C_{o}$ denote the input and output channel numbers of a
$K_H\times K_W$ sized 2d convolution kernel.
$\mathbf{X} \in \mathbb{R}^{C_i\times H\times W}$, 
$\mathbf{Y} \in \mathbb{R}^{C_{o}\times H'\times W'}$ 
denote the input and output tensors.
We omit the bias here as a common practice, the convolution process is denoted by
\begin{equation}
	\mathbf{Y} = \mathbf{W} \ast \mathbf{X}.
\end{equation}
%

\paragraph{Simplify a sequential structure.}
Consider a stack of convolutional layers denoted by
\begin{equation}
\mathbf{Y} = \mathbf{W}_N (\mathbf{W}_{N-1} \ast \cdots (\mathbf{W}_2 \ast (\mathbf{W}_1 \ast \mathbf{X}))),
\end{equation}
where $\mathbf{W}_j \in \mathbb{R}^{C_{j}\times C_{j-1}\times K_{H_j}\times K_{W_j}}$ 
satisfies $C_0=C_{i}, \;C_N=C_{o}$.
According to the associative law, such layers can be squeezed into one by convolving the kernels first
according to~Eq.~\eqref{eq:seq}.
\begin{equation}
\begin{split}
	\mathbf{Y} &= (\mathbf{W}_N (\mathbf{W}_{N-1} \ast \cdots (\mathbf{W}_2 \ast \mathbf{W}_1)) \ast \mathbf{X} \\
	&= \mathbf{W}_e \ast \mathbf{X},
\end{split}
\label{eq:seq}
\end{equation}
where $\mathbf{W}_j$ is the weight of the $j^{th}$ layer. 
$\mathbf{W}_e$ denotes the end-to-end mapping matrix. 
The pixel-wise form of Eq.~\eqref{eq:seq} 
is shown in the supplementary materials.



\paragraph{Simplify a parallel structure.}
The simplification of a parallel structure is trivial.
According to the linearity of convolution, 
we can merge multiple branches into one according to Eq.~\eqref{eq:merge-paral}.
\begin{equation}
	\mathbf{Y} = \sum_{m=0}^{M-1} (\mathbf{W}_m \ast \mathbf{X}) = (\sum_{m=0}^{M-1} \mathbf{W}_m) \ast \mathbf{X},
\label{eq:merge-paral}
\end{equation}
where $\mathbf{W}_m$ is the weight of the $m^{th}$ branch, 
and  $(\sum_{m=0}^{M-1} \mathbf{W}_m)$ is the unified weight. 
It is worth noting that when merging kernels with different sizes, 
we need to align the spatial centers of different kernels, \eg,
an 1$\times$1 kernel should be aligned with the center of a 3$\times$3 kernel.

\paragraph{Training overhead: from features to kernels.}
No matter how complex the block is, 
it must be constituted by no more than multi-branch and multi-layer sub-topologies. 
Thus, it can be simplified into a single one according to the two simplification rules above. 
Finally, we could get the all-in-one end-to-end mapping weight 
and only convolve once during training.
According to Eq.~\eqref{eq:seq} and Eq.~\eqref{eq:merge-paral}, 
we actually convert the operations~(convolution, addition) on intermediate 
feature maps into those on convolutional kernels.
As a result, we reduce the extra training cost of a re-param block
from $O(H\times W)$ to $O(K_H\times K_W)$.

\subsection{Gradient Analysis on Multi-branch Topology}
\label{sec:principle} To understand why the block linearization step is feasible, \ie why the scaling layers are important, we conduct analysis on the optimization of the unified weight re-parameterized. Our conclusion is that for the branches with norm layers removed, 
the utilization of scaling layers could diversify their optimization directions, and prevent them from degrading into a single one.

To simplify the notation, we take only single dimension of the output $\mathbf{Y}$. Consider a conv-scaling sequence (a simplified version of conv-norm sequence):
\begin{equation}
\Phi^{Conv-Scale} := \{\mathbf{y} = \gamma \mathbf{W}\mathbf{x}|W \in \mathbb{R}^{o,i}, \gamma \in \mathbb{R}^{o}\},
\end{equation}
where $I=C_i\times K_H\times K_W$, $\mathbf{x} \in \mathbb{R}^I$ is vectorized pixels inside a sliding window,  $y \in \mathbb{R}^O$, $O=1$, $\mathbf{W}$ is a convolutional kernel corresponding to certain output channel, and $\gamma$ is the scaling factor. Suppose all parameters are updated by stochastic gradient descent, the mapping $\mathbf{W}_{cs} := \gamma \mathbf{W}$ is updated by:

\begin{equation}
\label{eq:sgdcn}
\begin{split}
&\mathbf{W}^{(t+1)}_{cs} :=  \gamma^{(t+1)} \mathbf{W}^{(t+1)} \\ &= (\gamma^{(t)}-\eta\mathbf{W}^{(t)}\mathbf{x}^{\top}\frac{\partial L}{\partial\mathbf{y}})(\mathbf{W}^{(t)}-\eta\gamma^{(t)}\mathbf{x}^{\top}\frac{\partial L}{\partial\mathbf{y}}) \\ &= 
\mathbf{W}^{(t)}_{cs} - \eta (vec(diag(\mathbf{W}^{(t)})^{2}) +\left \| \gamma^{(t)} \right \|_2^2)\mathbf{x}^{\top}\frac{\partial L}{\partial\mathbf{y}}
\end{split}
\end{equation}
where $L$ is the loss function of the entire model and $\eta$ is the learning rate. For a multi-branch topology with a shared $\gamma$, \ie:
\begin{equation}
\Phi^{Conv-Scale}_M := \{\mathbf{y} = \gamma \sum_{j=1}^{M}\mathbf{W}_{j}\mathbf{x}|\mathbf{W}_{j} \in \mathbb{R}^{o,i}, \gamma \in \mathbb{R}^{o}\},
\end{equation}
the end-to-end weight $\mathbf{W}_{e_1,cs} := \gamma\sum_{j=1}^{M}\mathbf{W}_{j}$ is optimized equally from that of Eq. \eqref{eq:sgdcn}:
\begin{equation}
\mathbf{W}^{(t+1)}_{e_1,cs} -  \mathbf{W}^{(t)}_{e_1,cs} = \mathbf{W}^{(t+1)}_{cs} -  \mathbf{W}^{(t)}_{cs},
\end{equation}
with the same forwarding $t^{th}$-moment end-to-end matrix $\mathbf{W}^{(t)}_{cs} = \mathbf{W}^{(t)}_{e_1,cs}$.
Hence, $\Phi^{Conv-Scale}_M$ introduces no optimization change. This conclusion is also supported experimentally \cite{Ding21dbb}. On the contrary, a multi-branch topology with branch-wise $\gamma$ provide such changes, \eg:
\begin{equation}
\Phi^{Conv-Scale}_{M,2} := \{\mathbf{y} = \sum_{j=1}^{M}\gamma_{j}\mathbf{W}_{j}\mathbf{x}|\mathbf{W}_{j} \in \mathbb{R}^{o,i}, \gamma_{j} \in \mathbb{R}^{o}\}.
\end{equation}
The end-to-end weight $\mathbf{W}_{e_2,cs} := \sum_{j=1}^{M}\gamma_{j}\mathbf{W}_{j}$ is updated differently from that of Eq. \eqref{eq:sgdcn}:
\begin{equation}
\label{eq:nlsp}
\mathbf{W}^{(t+1)}_{e_2,cs} -  \mathbf{W}^{(t)}_{e_2,cs} \ne \mathbf{W}^{(t+1)}_{cs} -  \mathbf{W}^{(t)}_{cs},
\end{equation}
with the same precondition $\mathbf{W}^{(t)}_{cs} = \mathbf{W}^{(t)}_{e_2,cs}$ and Condition \ref{cond:act} satisfied:

\begin{condition} \label{cond:act}
	At least two of all the branches are active.
	\begin{equation}
	\begin{split}
	\exists \;& \mathbf{S} \subseteq \{1, 2, \cdots, M\}, \left | \mathbf{S}  \right | \ge 2, \\ &such \; that \; \forall j \in \mathbf{S} ,\;   vec(diag(\mathbf{W}_j^{(t)})^{2}) +\left \| \gamma_j^{(t)} \right \|_2^2 \ne \mathbf{0}
	\end{split}
	\end{equation}	
\end{condition}

\begin{condition} \label{cond:ne}
	The initial state of each active branch is different from that of each other.
	\begin{equation}
	\forall j_1, j_2 \in \mathbf{S}, \; j_1 \ne j_2, \quad	\mathbf{W}_{j_1}^{(0)} \ne \mathbf{W}_{j_2}^{(0)}.
	\end{equation}	
\end{condition}

Meanwhile, when Condition \ref{cond:ne} is met, the multi-branch structure will not degrade into single one for both forwarding and backwarding. This reveals the following proposition explaining why the scaling factors are important.  Note that both Condition \ref{cond:act} and \ref{cond:ne} are always met when weights $\mathbf{W}_{j}^{(0)}$ of each branch is random initialized \cite{he15} and scaling factors $\gamma_{j}^{(0)}$ are initialized to 1. 

\begin{proposition}
	\label{prop}
	A single-branch linear mapping, when re-parameterizing parts or all of it by over-two-layer multi-branch topologies, the entire end-to-end weight matrix will be differently optimized. If one layer of the mapping is re-parameterized to up-to-one-layer multi-branch topologies, the optimization will remain unchanged.
\end{proposition} 



So far, we have extended the discussion on how re-parameterization impacts optimization, from multi-layer only \cite{Arora18} to multi-branch included as well. Actually, all current effective re-parameterization topology \cite{Ding19, Ding21dbb, Ding21repvgg, Guo20, Cao20} can be validated by either \cite{Arora18} or Proposition \ref{prop}. For detailed derivation and discussion about this subsection, please refer to supplementary materials.

\subsection{Block Design}
\label{sec:components}
\begin{figure}[tbp]
	\begin{center}
		\includegraphics[width=1.0\linewidth]{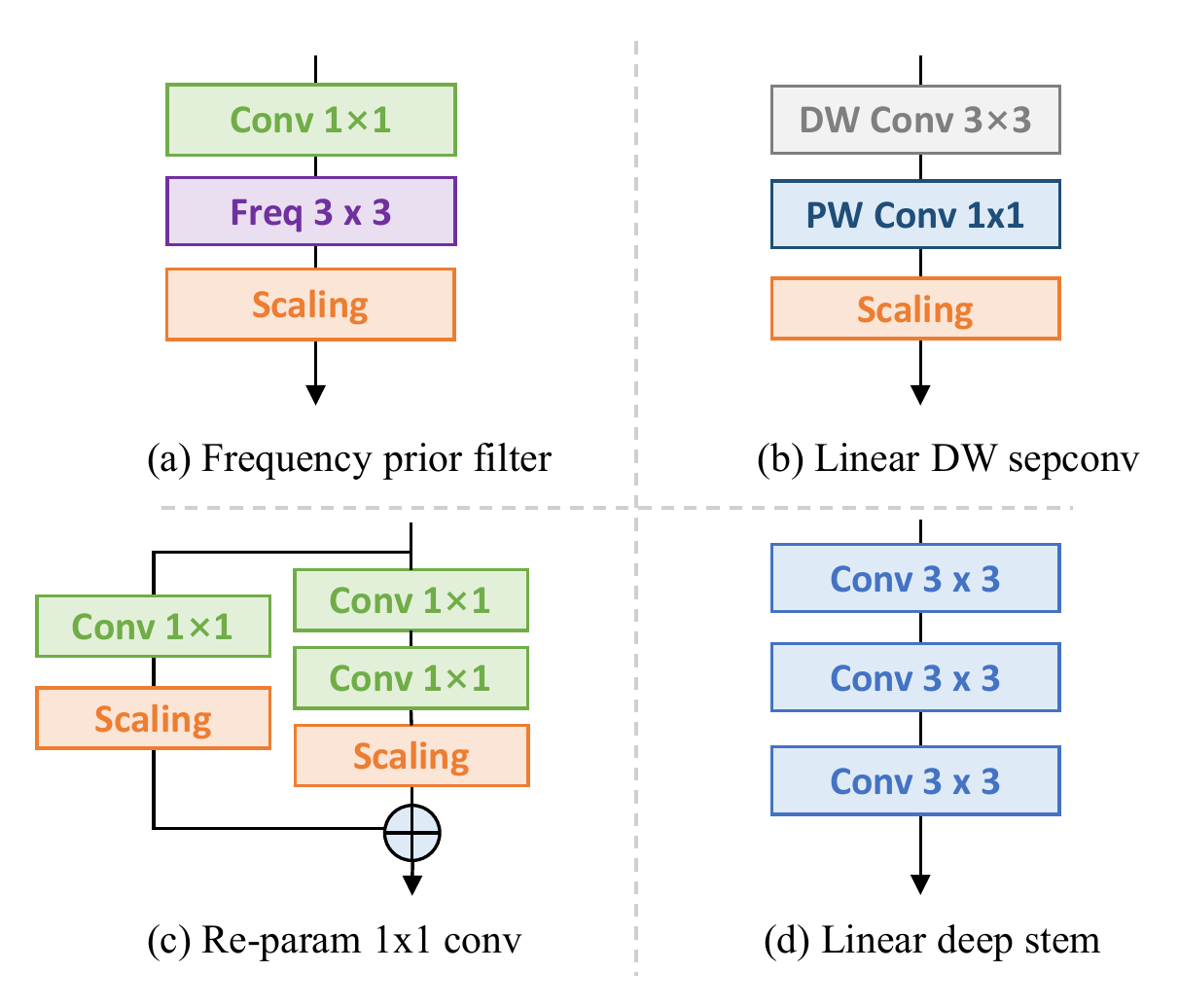}
	\end{center}
	\vspace{-.7em}
	\caption{Illustration of the proposed four components in \secref{sec:components}.}
	\label{fig:components}
\end{figure}

Since the proposed OREPA saves the training cost by a large margin,
it enables us to explore more complicated training blocks.
Hereby, We design a novel of re-parameterization models,
\ie, OREPA-ResNet, by 
linearizing the state-of-the-art model~DBB~\cite{Ding21dbb}, 
and inserting the following components~(\figref{fig:components}).

\vspace{-1.25em}
\paragraph{Frequency prior filter.}
In previous work~\cite{Ding21dbb}, pooling layers are utilized in the block.
According to Qin~\etal~\cite{Qin21}, the pooling layer is a special case of the frequency filters.
To this end, we add a Conv1$\times$1 - Frequency Filter branch.

\vspace{-1.25em}
\paragraph{Linear depthwise separable convolution. \cite{Cao20}} 
We slightly modify the depthwise separable convolution~\cite{Chollet17}
by removing the intermediate non-linear activation layer, 
making it feasible to be merged during training. 

\vspace{-1.25em}
\paragraph{Re-parameterization for 1$\times$1 convolution.}
Previous works mainly focus on the Re-parameter for 3$\times$3 convolutional layers but ignore
the 1$\times$1 ones.
We propose to re-parameterize 1$\times$1 layers 
because they play an important role in the bottleneck structures~\cite{He16, Brock21nfnet}.
Specifically, we add an additional Conv1$\times$1 - Conv1$\times$1 branch.

\vspace{-1.25em}
\paragraph{Linear deep stem.}
Large convolutional kernels are usually placed in the very beginning layers, \eg, 7$\times$7 stem layers~\cite{He16}, 
aiming at achieving a larger receptive field. 
Guo \etal replace the 7$\times$7 conv layer with stacked 3$\times$3 layers for a higher accuracy~\cite{Guo20}.
However, the stacked convs at the very beginning layers require larger computational overhead 
due to the high-resolution.
Note that we can squeeze the stacked deep stem with our proposed linear scaling layers
to a 7$\times$7 conv layer, 
which can greatly reduce the training cost while keep the high accuracy.


%
\begin{figure}[tbp]
	\begin{center}
		\includegraphics[width=1.0\linewidth]{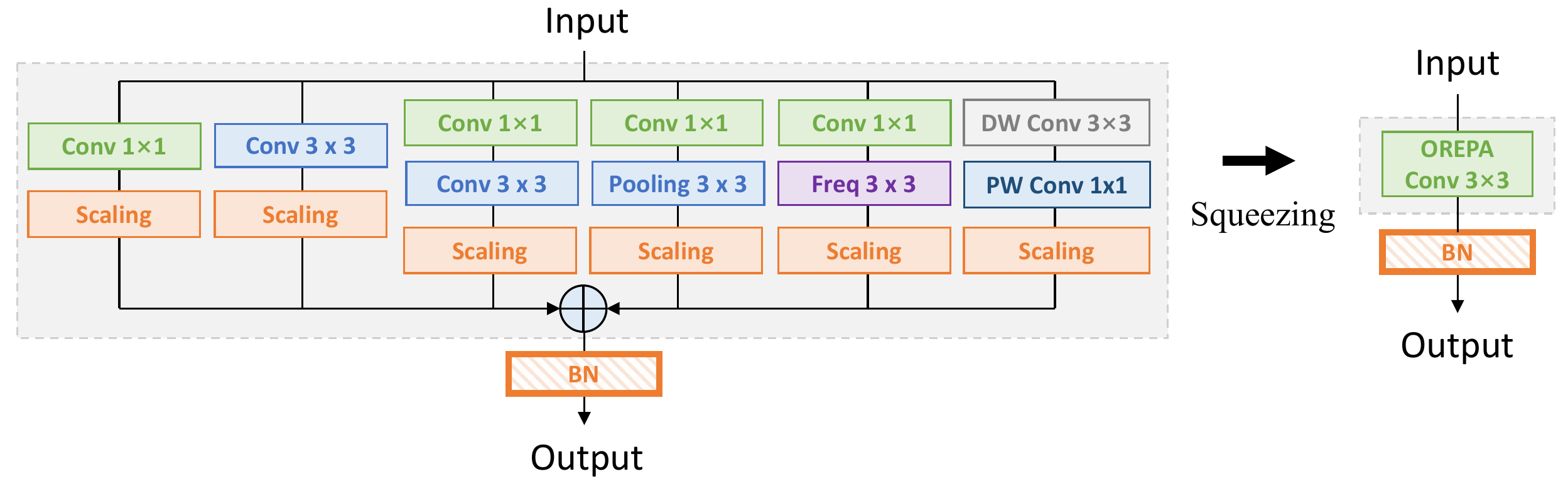}
	\end{center}
	\vspace{-.7em}
	\caption{The design of the proposed OREPA block, corresponding to a 3$\times$3 convolution
	during training and inference.}
	\label{fig:blocks}
\end{figure}
For each block in the OREPA-ResNet~(\figref{fig:blocks}),
i) we add a \emph{frequency prior filter} 
and a \emph{linear depthwise seperable convolution}. 
ii) We replace all the stem layers~(\ie, the initial 7$\times$7 convolution) with 
the proposed \emph{linear deep stem}.
iii) In the bottleneck~\cite{He16} blocks, in addition to i) and ii), 
we further replace the original 1$\times$1 convolution branch with the 
proposed Rep-1$\times$1 block. 


%% file: sections/experiment.tex
\section{Experimental Results}
\label{sec:exp}


\subsection{Implementation Details}
\label{sec:imp}

We conduct experiments on the ImageNet-1k~\cite{Deng09} dataset.
We follow the standard data pre-processing pipeline,
including random cropping, resizing~(to 224$\times$224), random horizontal flipping, and normalization. 
By default, we apply an SGD optimizer to train the models, 
with initial learning rate 0.1 and cosine annealed in 120 epochs.
We also linearly warm up the learning rate~\cite{He19} in the initial 5 epochs.
We use a global batch size 256 on 4 Nvidia Tesla V100 (32G) GPUs.
Unless specified, we use \textbf{\emph{ResNet-18}} as the base structure. 
For a fair comparison, we report results of different models trained 
under the same settings as described above.
For more details, please refer to our supplementary materials.



\subsection{Ablation Study}
\label{sec:ablation}
\begin{table}[tbp]
	\caption{
	Effectiveness of different components in OREPA. 
	We use Top-1 Accuracy to measure the performance.
	Note that Rep-1$\times$1 is design for the bottleneck blocks, thus not used in ResNet-18.
	}
	\vspace{1.0em}
	\small
	\centering\
	\begin{tabular}{lcc}
		\toprule
		  Model 								& ResNet-18 & ResNet-50 \\ \midrule
											
		  Baseline~\cite{He16}					& 71.21 	& 76.70 	\\
		  + Offline DBB blocks~\cite{Ding21dbb}	& 71.77		& 76.89		\\ \midrule
		+ Online								& 71.75		& 76.86		\\ 
		+ Frequency prior filter				& 71.78		& 76.92 	\\ 
		+ Linear depthwise separable			& 71.82		& 76.98 	\\
		+ Linear deep stem						& 72.13 	& 77.09 	\\
		+ Rep-1$\times$1								& - 		& 77.31 	\\ \bottomrule
	\end{tabular}
	\label{tab:components}
\end{table}
\begin{figure}[tbp]
	\begin{center}
		\includegraphics[width=1.0\linewidth]{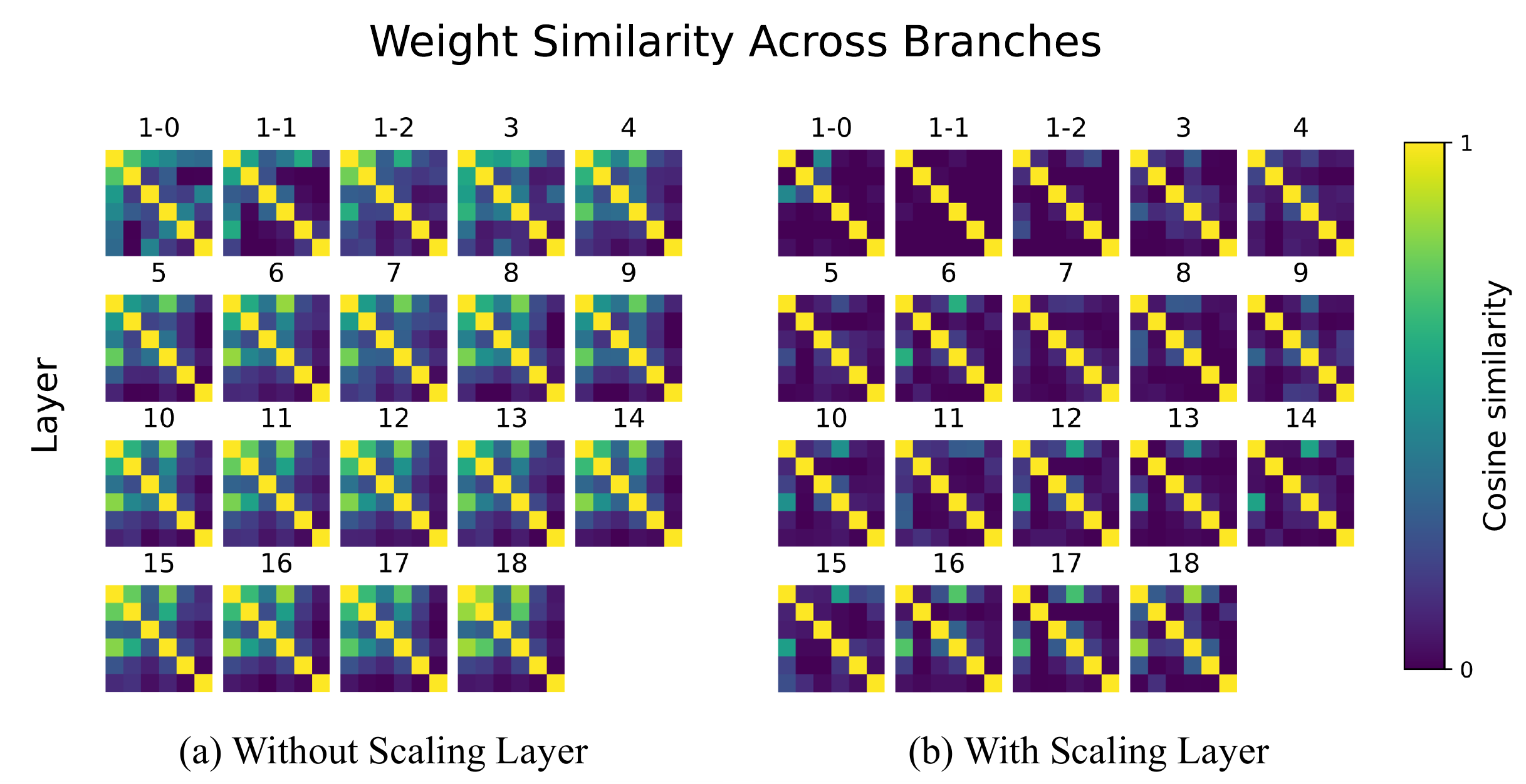}
	\end{center}
	\vspace{-1.0em}
	\caption{Visualization of branch-level similarity. 
	We calculate cosine similarities between the weights from different branches.
	}
	\label{fig:viz}
\end{figure}
\paragraph{Linear scaling layer and optimization diversity.}
We first conduct experiments to validate our core idea, that the proposed linear scaling layers
play a similar role as the normalization layers.
According to the analysis in \secref{sec:principle}, we show both the scaling layers and 
the norm layers are able to diversify the optimization direction.
To verify this, we visualize the branch-wise similarity
of all the branches in \figref{fig:viz}.
We find that the use of scaling layer can significantly increase the diversity of different branches.

We validate the effectiveness of such diversity in \tabref{tab:components}.
Take the ResNet-18 structure as an example, the two kinds of layers~(norm and linear scaling) bring similar 
performance gain~(\ie, 0.42 \vs 0.40).
This strongly supports our claim that 
it is the \textbf{\emph{scaling}} part, rather than the statistical-normalization part, 
that counts most in re-parameterization.

\vspace{-1.25em}
\paragraph{Various linearization strategies.}
\begin{table}[tb]
	\caption{
	Comparison with linearization variants on the ResNet-18 model. 
	Note that ``NaN'' means the gradient explosion 
	and the model fails to converge.}
	\vspace{.6em}
	\small
	\centering
	\begin{tabular}{@{}lc@{}}
		\toprule
		Linearization Variants 		& Top1-Accuracy(\%) \\ \midrule
		Vector scaling 				& 72.13 			\\
		Scalar scaling 				& 72.04				\\
		W/o scaling 				& 71.87				\\
		W/o post-addition norm 		&  NaN	 			\\ \bottomrule
	\end{tabular}
	\label{tab:linearization}
\end{table}
We attempt various linearization strategies for the scaling layer.
Specifically, we visit four variants. 
\begin{itemize}
	\vspace{-.5em}
	\item 
	Vector: utilize a channel-wise vector and perform the scaling operation along the channel axis.
	\vspace{-.5em}
	\item 
	Scalar: scale the whole feature map with a scalar.
	\vspace{-.5em}
	\item 
	W/o scaling: remove the branch-wise scaling layers.
	\vspace{-.5em}
	\item 
	W/o post-addition norm: remove the post-addition norm layer.
\end{itemize}
\vspace{-.5em}
From Table \ref{tab:linearization} we find that 
deploying either scalar scaling layers or no scaling layers leads to inferior results. 
Thus, we choose the vector scaling as the default strategy. 

We also study the effectiveness of the post-addition norm layers.
As stated in \secref{sec:linearization}, we add such layers to 
stabilize the training process. 
To demonstrate this, we remove such layers, as shown in the last row in 
\tabref{tab:linearization},
the gradients become infinity and the model fails to converge.

\vspace{-1.25em}
\paragraph{Each component matters.} 
Next, we demonstrate the effectiveness of the proposed components~\secref{sec:components}.
We conduct experiments on both the structures of both ResNet-18 and ResNet-50. 
As shown in Table \ref{tab:components},
each of the components helps to improve the performance. 

\vspace{-1.25em}
\paragraph{Online \vs offline.} 
\begin{figure}[tbp]
	\begin{center}
		\includegraphics[width=1.0\linewidth]{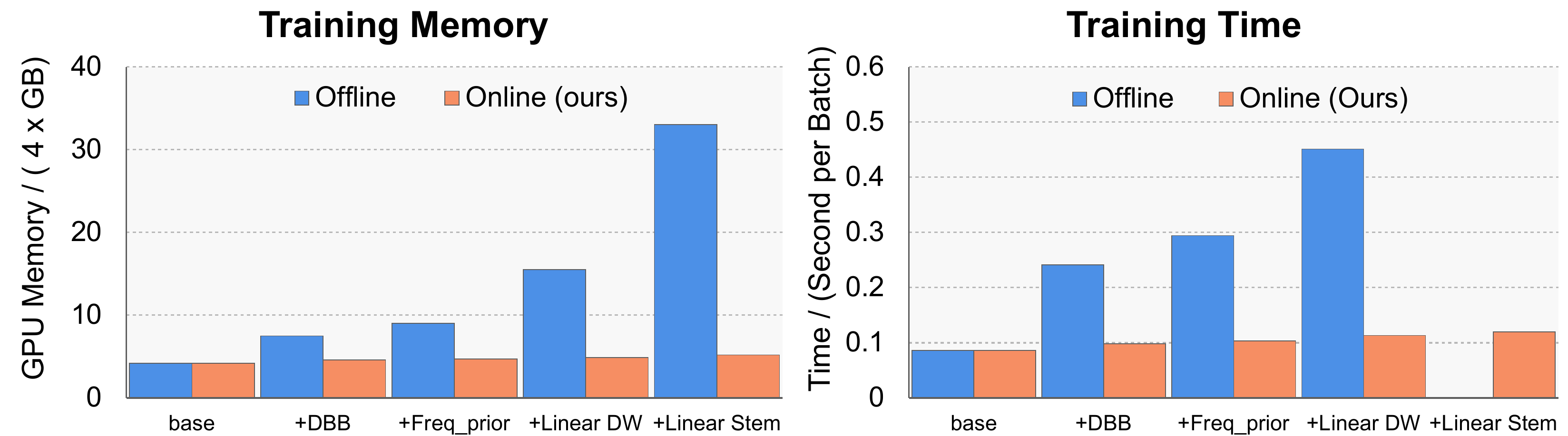}
	\end{center}
	\vspace{-.5em}
	\caption{Comparison of training cost between offline re-param method and OREPA.}
	\label{fig:budgets}
\end{figure}
%
We compare the training cost of 
OREPA-ResNet-18 with its offline counterpart~(\ie, DBB-18). 
We illustrate both the consumed memory~(\figref{fig:budgets}(a)) and the training time~(\figref{fig:budgets}(b)).
With the increased number of components, 
the offline re-parameterized model suffers from rapidly increasing memory utilization and the long training time. 
We can not even introduce deep stem to a ResNet-18 model due to the high memory cost. 
In contrast, the online re-param strategy accelerates the training speed by 4$\times$, and saves up to 96$+\%$ extra GPU memory.
The overall training overhead roughly lies in the same level as that of the base model~(vanilla ResNet).

\subsection{Comparison with other Re-Param methods}
\label{sec:cmp}
\begin{table}[tbp]
	\caption{Comparison with other re-parameterization models. 
	Note that instead of directly quote results from the original papers, we report models trained
	on our machine for a fairer comparison.}
	\vspace{.5em}
	\small
	\centering\
	\begin{tabular}{lcccc}
		\hline
		\toprule
		\multirow{2}{*}{Model} & 
		\multirow{2}{*}{Re-param} & 
		\multirow{1}{*}{Top1-} & 
		\multirow{1}{*}{GPU-} & 
		\multirow{1}{*}{Training} \\
		&  & Acc & Mem & time/batch \\
		
		\midrule
		
		\multirow{3}{*}{ResNet-18}
		& None & 71.21 & 4.2G & 0.086s \\
		& DBB & 71.77 & 7.5G  & 0.241s \\
		& OREPA & 72.13 & 5.2G & 0.120s \\  
		\midrule

		\multirow{3}{*}{ResNet-34}
		& None & 74.13 & 5.2G & 0.116s \\
		& DBB & 74.83 & 11.1G & 0.424s \\
		& OREPA & 75.04 & 6.7G & 0.182s \\ 
		\midrule  
		
		\multirow{3}{*}{ResNet-50}
		& None & 76.70 & 10.3G & 0.204s \\
		& DBB & 76.89 & 13.7G & 0.380s \\
		& OREPA & 77.31 & 11.2G & 0.252s \\ 
		\midrule
		 
		\multirow{3}{*}{ResNet-101}
		& None & 77.71 & 14.5G & 0.382s \\
		& DBB & 78.02 & 19.7G& 0.641s \\
		& OREPA & 78.29 & 16.1G& 0.445s \\ \midrule \midrule
		
		
		\multirow{2}{*}{RepVGG-A0}
		& RepVGG & 72.41 & 3.8G & 0.100s \\
		& OREPA & 73.04 & 4.2G & 0.136s \\ 
		\midrule
		
		\multirow{2}{*}{RepVGG-A1}
		& RepVGG & 74.46 & 4.8G & 0.121s \\
		& OREPA & 74.85 & 5.5G & 0.147s \\ 
		\midrule
		
		\multirow{2}{*}{RepVGG-A2}
		& RepVGG & 76.48 & 5.8G & 0.170s \\
		& OREPA & 76.72 & 7.4G & 0.210s \\ \midrule	\midrule
		
	\end{tabular}
	\label{tab:res}
\end{table}

We compare our methods with previous structural re-parameterization methods on ImageNet.
For the ResNet structure, we compare OREPA-ResNet series with DBB~\cite{Ding21dbb}.
\emph{For a fairer comparison, we report results of DBB trained using our setting.} 
This makes the performances slightly higher than those reported in~\cite{Ding21dbb}.

From \tabref{tab:res}, we observe that on the ResNet-series, 
OREPA can consistently improve performances on various models by up to +0.36$\%$.
At the same time, it accelerates the training speed by 1.5$\times$ to 2.3$\times$ 
and saves around 70+$\%$ extra training-time memory caused by re-param. 
We also conduct experiments on the VGG structure, we compare OREPA-VGG with RepVGG~\cite{Ding21repvgg}. 
For the OREPA-VGG models, we simply replace the Conv-3$\times$3 branch with the one we used in OREPA-ResNet.
Such a modification only introduces marginal extra training cost, 
while brings clear performance gain~(+0.25\%$\sim$+0.6\%).


\subsection{Object Detection and Semantic Segmentation}
\label{sec:detseg}
To validate the generalization of online re-parameterized models, 
we apply the pretrained OREPA-ResNet-50 to conduct experiments on the object detection and semantic segmentation tasks. 
On MS-COCO~\cite{Lin14}, we train the commonly-used detectors, \ie, Faster RCNN~\cite{Ren15} and RetinaNet~\cite{Lin17} 
with default settings in mmdetection~\cite{mmdetection} for 12 epochs. 
On Cityscapes, 
we train the PSPNets~\cite{Zhao17} and DeepLabV3+~\cite{Chen18} models using mmsegmentation \cite{mmseg2020} for 40K steps.
As show in Table \ref{tab:detseg}, OREPA consistently improves performances on the two tasks.


\begin{table}[tbp]
	\caption{
	Performance on COCO and Cityscapes validation set with pretrained ResNet-50 backbones.
	``Mem'' is the GPU memory~(GB) cost during training and ``Time'' denotes the 
	average training time of a step (minutes per 50 steps).
	}
	\vspace{.6em}
	\label{tab:detseg}
	\centering
	\small
	\begin{tabular}{lccc|ccc}
		\toprule
		\multirow{2}{*}{Re-param} & \multicolumn{3}{c}{Faster-RCNN} & \multicolumn{3}{c}{RetinaNet} \\ \cmidrule{2-7} 
				& mAP 	& Mem 	& Time 	& mAP & Mem & Time \\ \midrule
		None 	& 36.5 	& 5.0	& 0.37	& 36.2 & 5.0	& 0.29	\\
		DBB 	& 36.5 	& 6.5	& 0.46	& 36.6 & 6.5	& 0.40	 \\
		OREPA 	& 37.0 	& 5.7	& 0.39 	& 36.9 & 5.9	& 0.36	 \\ \midrule \midrule
		\multirow{2}{*}{Re-param} & \multicolumn{3}{c}{PSPNet} & \multicolumn{3}{c}{DeepLabV3+} \\ \cmidrule{2-7} 
				& mIoU 	& Mem 	& Time 	& mIoU & Mem & Time \\ \midrule
		None 	& 74.47	& 11.5 	& 0.31	& 76.63 & 13.1  & 0.50\\
		DBB 	& 74.50	& 13.0 	& 0.54	& 77.15 & 14.6 	& 0.71\\
		OREPA 	& 75.47	& 12.3 	& 0.41	& 77.59 & 13.8	& 0.56\\ \bottomrule
	\end{tabular}
\end{table}

\subsection{Limitations}
When simply transferring the proposed OREPA from ResNet to RepVGG, 
we find inconsistent performances between the residual-based and residual-free~(VGG-like) structures. 
Therefore, 
we reserve all the three branches in the RepVGG block 
to maintain a competitive accuracy, 
which brings marginally increased computational cost. 
This is an interesting phenomenon, 
and we will briefly discuss this in the supplementary materials.

%% file: sections/conclusion.tex
\vspace{-0.35em}\section{Conclusion}
In this paper, we present online convolutional re-parameterization (OREPA), a two-stage pipeline aiming to reduce the huge training overhead by squeezing the complex training-time block into a single convolution. To achieve this goal, we replace the training-time non-linear norm layers with linear scaling layers, which maintains optimization diversity and the enhancement of the representational capacity. As a result, we significantly reduce the training-budgets for re-parameterization models. This is essential for training large-scale neural network with complex topologies, and it further allows us to re-parameterize models in a more economic and effective way.
Results on various tasks demonstrate the effectiveness of OREPA 
in terms of both accuracy and efficiency.

\vspace{-0.5em}\paragraph{Acknowledgments.} This work was supported by the National Key R\&D Program of China under Grant 2020AAA0103901, as well as Key R\&D Development Plan of Zhejiang Province under Grant 2021C01196. We thank Dongxu Wei and Hualiang Wang for valuable suggestions.

%% file: sections_supp/method.tex
 \section{Online Re-Parameterization}

\subsection{Block Squeezing: More Details}
\paragraph{Pixel-wise definition of convolution.} Let $C_{i}$, $C_{o}$ denote the input and output channel numbers of a $K_H\times K_W$ sized 2d convolution kernel.
$\mathbf{X} \in \mathbb{R}^{C_i\times H\times W}$ and
$\mathbf{Y} \in \mathbb{R}^{C_{o}\times H'\times W'}$ 
denote the input and output tensors. The pixel-wise form of convolution $\mathbf{Y} = \mathbf{W} \ast \mathbf{X}$ is:
\begin{equation}
\mathbf{Y}_{c_{o},h,w} = \sum_{c_{i}=0}^{C_{i}-1}\sum_{k_h=0}^{K_H-1} \sum_{k_w=0}^{K_W-1} \mathbf{W}_{c_{i},c_{o},k_{h},k_{w}}\mathbf{X}_{c_{i},h+\Delta k_{h},w+\Delta k_{w}},
\end{equation}
where $\Delta k_{h} = k_{h} - \left \lfloor \frac{K_H-1}{2}  \right \rfloor$ and $\Delta k_{w} = k_{w} - \left \lfloor \frac{K_W-1}{2}  \right \rfloor$. Similarly, the pixel-wise form of convolution between two kernels 
$\mathbf{W}_{j,j+1} = \mathbf{W}_{j+1} \ast \mathbf{W}_j$ is defined in Eq.~\eqref{eq:merge-seq}.

\begin{equation}
\begin{split}
\mathbf{W}_{j,j+1_{c_{q}, c_{p},u,v}} &= \sum_{c_{j}=0}^{C_{j}-1}\sum_{k_h=0}^{K_{H_j}-1} \sum_{k_w=0}^{K_{W_j}-1} \mathbf{W}_{{j+1}_{c_{q},c_{j},k_{h},k_{w}}}\\&\cdot \mathrm{Pad}(\mathbf{{W}}_{j}, {K_{W_{j+1}}-1}, {K_{W_{j+1}}-1},  {K_{H_{j+1}}-1}, {K_{H_{j+1}}-1} )_{c_{j},c_{p},u-\Delta k_{h},v-\Delta k_{w}}\\ \textrm{or} &\quad\quad\\\mathbf{W}_{j,j+1_{c_{q}, c_{p},u,v}} &= \sum_{c_{j}=0}^{C_{j}-1}\sum_{k_h=0}^{K_{H_{j+1}}-1} \sum_{k_w=0}^{K_{W_{j+1}}-1} \mathbf{W}_{{j}_{\;c_{j},c_{p},k_{h},k_{w}}}\\&\cdot \mathrm{Pad}(\mathbf{{W}}_{j+1}, {K_{W_{j}}-1}, {K_{W_{j}}-1},  {K_{H_{j}}-1}, {K_{H_{j}}-1} )_{c_{q},c_{j},u-\Delta k_{h},v-\Delta k_{w}},
\end{split}
\label{eq:merge-seq}
\end{equation}
where $\mathbf{W}_j \in \mathbb{R}^{C_{j}\times C_{j-1}\times K_{H_j}\times K_{W_j}}$ and $\mathbf{W}_{j+1} \in \mathbb{R}^{C_{j+1}\times C_{j}\times K_{H_{j+1}}\times K_{W_{j+1}}}$ are the weights of two sequential convolutional layers, and $\mathrm{Pad(\cdot, L, R, T, B)}$ 
means zero padding the weight tensor spatially 
from the left, right, top, and bottom by L, R, T, and B pixels, respectively. 
The inter-weight convolution is very similar to regular convolution, 
except that the order of element convolved is inverse (note the minus signs in the indices of $\mathbf{W}_j$ or $\mathbf{W}_{j+1}$).

\begin{table}[htbp]
	\centering
	\caption{Training-time linear layers deployed in the OPERA block. Note that some of the conv $1\times1$ layers in the block are randomly initialized, while other are identically initialized.}
	\vspace{0.4em}
	\renewcommand\arraystretch{1.5}
	\begin{tabular}{@{}ccccc@{}}
		\toprule
		Layer & Dimension of Weights & Initialization & Fixed & Comments \\ \midrule
		Conv & $\mathbb{R}^{C_{o}\times \left \lfloor \frac{C_{i}}{G}  \right \rfloor \times K_{H}\times K_{W}}$ & $\mathbf{W}_{c_o,  \left \lfloor \frac{c_i}{g}  \right \rfloor,k_{h},k_{w}} \sim U(0, \Theta \frac{1}{\sqrt{\left \lfloor \frac{C_{i}}{G}  \right \rfloor \times K_{H}\times K_{W}}}) $ & \XSolidBrush & \multicolumn{1}{c}{\begin{tabular}[c]{@{}c@{}}std conv, group conv,\\ and dw conv \small{($G=C_i$)} \end{tabular}} \\ $\textrm{Conv}_{1\times1}^\textrm{iden.}$
		& $\mathbb{R}^{C_{o}\times \left \lfloor \frac{C_{i}}{G}  \right \rfloor \times 1\times 1}$ & $\mathbf{W}_{c_o,  \left \lfloor \frac{c_i}{g}  \right \rfloor,1,1} \left\{\begin{matrix}
		1, \textrm{if}  \;  \frac{c_o}{C_o}=\frac{c_{i}}{C_{i}}  \\0,  \;   \textrm{else}
		\end{matrix}\right. $ & \XSolidBrush & init as an identity layer \\
		Scaling & $\mathbb{R}^{C}$ & $\mathbf{W}_c=m, m \in [0,1]$ & \CheckmarkBold & channel-wise scaling \\
		Pooling & $\mathbb{R}^{K_{H}\times K_{W}}$ &  $\mathbf{W}_{k_{h},k_{w}}=\frac{1}{K_{H}\times K_{W}}$ & \CheckmarkBold & avg pooling \\
		Filtering & $\mathbb{R}^{C\times K_{H}\times K_{W}}$ & $\mathbf{W}_{c,k_h,k_w}=\left\{\begin{matrix}\mathrm{cos}(\frac{(c+1)\times(k_h+0.5)\times\pi}{K_H}),c<\left \lfloor \frac{C}{2} \right \rfloor \\\mathrm{cos}(\frac{(c-\left \lfloor \frac{C}{2} \right \rfloor +1)\times(k_w+0.5)\times\pi}{K_W}),c\ge \left \lfloor \frac{C}{2} \right \rfloor 
		\end{matrix}\right.$ & \CheckmarkBold & freq prior filtering \\ \bottomrule
	\end{tabular}
	\label{tab:ll}
\end{table}

\vspace{-0.2em}
\paragraph{Efficiency analysis on parallel squeezing.} Here we discuss two cases deployed in our re-parameterization blocks. For the conv $1\times1$ - $k\times k$ sequences, the effectiveness of online squeezing has been discussed in Sec. 3.3 of the body. For re-parameterizing $k\times k$ conv into stacked $3\times 3$ convs, it can be controversial whether the online squeezing strategy should be applied since stacked $3\times 3$ convs are believed to be more resource friendly. However, we find it not the case for linear deep stem. This is because the intermediate feature maps are of high resolution, leading to large GPU utilization. Meanwhile, there are only three input channels for the stem layer, yet much more channels for the intermediate feature maps.

\vspace{-0.2em}
\paragraph{Degraded convolutions.} For the unification of the description of block squeezing, we regard all training-time linear layers as standard or degraded convolutional layers. The specific definition of various linear layers is listed in Table \ref{tab:ll}. Note that the weights of all listed layers can be homogeneously represented as weights of a corresponding convolutional kernel through repetitive extension on certain dimensions.

\subsection{Gradient Analysis on Multi-branch Topology: Detailed Derivations}

To understand why the block linearization step is feasible, \ie why the scaling layers are important, we conduct analysis on the optimization of the unified weight re-parameterized. Our conclusion is that for the branches with norm layers removed, 
the utilization of scaling layers could diversify their optimization directions, and prevent them from degrading into a single one.
Let us begin with a single-branch multi-layer topology, whose training dynamic has been theoretically discussed in \cite{Arora18}. To simplify the notation, we take only single dimension of the output $\mathbf{Y}$. And the convolutional layer is represented as a linear system:
\begin{equation}
\Phi := \{\mathbf{y} = \mathbf{W}\mathbf{x}|\mathbf{W} \in \mathbb{R}^{O\times I}\},
\end{equation}
where $I=C_i\times K_H\times K_W$, $\mathbf{x} \in \mathbb{R}^I$ is vectorized pixels inside a sliding window,  $y \in \mathbb{R}^O$, $O=1$, and $\mathbf{W}$ is a convolutional kernel corresponding to certain output channel. Suppose $\mathbf{W}$ optimizated by stochastic gradient descent with a learning rate $\eta$:
\begin{equation}
\label{eq:sgd}
\mathbf{W}^{(t+1)} := \mathbf{W}^{(t)} - \eta \mathbf{x}^{\top} \frac{\partial L}{\partial \mathbf{y}},
\end{equation}
where $L$ is the loss function of the entire model. Consider stacked multiplicative (\ie multi-layer) re-parameterization:
\begin{equation}
\label{eq:ml}
\Phi_{N} := \{\mathbf{y} = \mathbf{W}_{N}\mathbf{W}_{N-1}\cdots \mathbf{W}_{1}\mathbf{x}|\mathbf{W}_j \in \mathbb{R}^{n_j\times n_{j-1}}\}.
\end{equation}

\begin{lemma}
	\label{lemma} \cite{Arora18} 
	Since each component $\mathbf{W}_j$ is updated by stochastic gradient descent respectively, the end-to-end mapping matrix $\mathbf{W}_e := \mathbf{W}_{N}\mathbf{W}_{N-1}\cdots \mathbf{W}_{1}$ is optimized differently from that of Eqn. \eqref{eq:sgd}:
	\begin{equation}
	\begin{split}
	\label{eq:lemma} \mathbf{W}_{e}^{(t+1)} :&= \mathbf{W}_{N}^{(t+1)}\mathbf{W}_{N-1}^{(t+1)}\cdots\mathbf{W}_{1}^{(t+1)} \\ =  &\mathbf{W}_e^{(t)} -  \eta \left \| \mathbf{W}_e^{(t)} \right \|_2^{2-\frac{2}{N}}   	\cdot  (\mathbf{x}^{\top}\frac{\partial L}{\partial \mathbf{y}} + (N-1) \cdot Pr_{\mathbf{W}_e^{(t)}}\{\mathbf{x}^{\top}\frac{\partial L}{\partial \mathbf{y}}\}) + O(\eta^2),
	\end{split}
	\end{equation}
\end{lemma}
where 
\begin{equation}
\begin{split}
\label{eq:proj}
Pr_{\mathbf{W}}\{\mathbf{G}\} := \left\{\begin{matrix} \frac{\mathbf{W}}{\left \| \mathbf{W} \right \|_2 }\mathbf{G}^{\top}\cdot \frac{\mathbf{W}}{\left \| \mathbf{W} \right \|_2 } &, \mathbf{W} \ne 0 \\ 0 &, \mathbf{W} = 0 \end{matrix}\right.	\end{split}
\end{equation}
is the projection of $\mathbf{G}$ the direction of $\mathbf{W}$, and
\begin{equation}
	\mathbf{W}^{(t+1)}_i := \mathbf{W}_i^{(t)} - \eta\frac{\partial L}{\partial \mathbf{W}_i^{(t)}} = \mathbf{W}_i^{(t)} - \eta {(\prod_{j=N}^{i+1}\mathbf{W}_j^{(t)}) (\prod_{m=i-1}^{1}\mathbf{W}_m^{(t)} \mathbf{x})^{\top}}\frac{\partial L}{\partial \mathbf{y}}, \forall i \in \{1, 2, \cdots, M\}.
\end{equation}

From Lemma \ref{lemma} we can directly know that the update item of $\mathbf{W}_e$ is changed both in norm and direction, due to the multi-layer (at least 2-layer) topology. To further understand the optimization of multi-branch re-parameterization, consider a convolution-scaling sequence:
\begin{equation}
\Phi^{Conv-Scale} := \{\mathbf{y} = \gamma \mathbf{W}\mathbf{x}|W \in \mathbb{R}^{o,i}, \gamma \in \mathbb{R}^{o}\},
\end{equation}
where $\mathbf{W}$ and $\gamma$ are weights of the convolutional layer and the scaling layer respectively. The mapping $\mathbf{W}_{cs} := \gamma \mathbf{W}$ is updated by:
\begin{equation}
\label{eq:sgdcn}
\begin{split}
&\mathbf{W}^{(t+1)}_{cs} :=  \gamma^{(t+1)} \mathbf{W}^{(t+1)} \\ &= (\gamma^{(t)}-\eta\mathbf{W}^{(t)}\mathbf{x}^{\top}\frac{\partial L}{\partial \mathbf{y}})(\mathbf{W}^{(t)}-\eta\gamma^{(t)}\mathbf{x}^{\top}\frac{\partial L}{\partial \mathbf{y}}) \\ &= 
\mathbf{W}^{(t)}_{cs} - \eta (vec(diag(\mathbf{W}^{(t)})^{2}) +\left \| \gamma^{(t)} \right \|_2^2)\mathbf{x}^{\top}\frac{\partial L}{\partial \mathbf{y}} + O(\eta^2)
\end{split}
\end{equation}

Note that Eqn. \eqref{eq:sgdcn} is a special case of Lemma \ref{lemma}, corresponding to the fact that the conv-scale sequence is actually a two-layer topology. For a multi-branch topology with a shared $\gamma$, \ie:
\begin{equation}
\Phi^{Conv-Norm}_M := \{\mathbf{y} = \gamma \sum_{j=1}^{M}\mathbf{W}_{j}\mathbf{x}|\mathbf{W}_{j} \in \mathbb{R}^{o,i}, \gamma \in \mathbb{R}^{o}\},
\end{equation}
the end-to-end weight $\mathbf{W}_{e_1,cs} := \gamma\sum_{j=1}^{M}\mathbf{W}_{j}$ is optimized equally from that of Eqn. \eqref{eq:sgdcn}:
\begin{equation}
\begin{split}
\mathbf{W}^{(t+1)}_{e_1,cs} :&=  \mathbf{W}^{(t)}_{e_1,cs} - \eta (vec(diag(\sum_{j=1}^{M} \mathbf{W}_j^{(t)})^{2}) +\left \| \gamma^{(t)} \right \|_2^2)\mathbf{x}^{\top}\frac{\partial L}{\partial \mathbf{y}} \\ &= \mathbf{W}^{(t)}_{e_1,cs} - \eta (vec(diag(\mathbf{W}^{(t)})^{2}) +\left \| \gamma^{(t)} \right \|_2^2)\mathbf{x}^{\top}\frac{\partial L}{\partial \mathbf{y}} + O(\eta^2),
\end{split}
\end{equation}
with the same forwarding $t^{th}$-moment end-to-end matrix $\mathbf{W}^{(t)}_{cs} = \mathbf{W}^{(t)}_{e,cs}$, which equivalently means $\sum_{j=1}^{M} \mathbf{W}_j^{(t)} = \mathbf{W}^{(t)}$.
Hence, $\Phi^{Conv-Scale}_M$ introduces no optimization change. This conclusion is also supported experimentally \cite{Ding21dbb}. On the contrary, a multi-branch topology with branch-wise $\gamma$ provide such changes, \eg:
\begin{equation}
\Phi^{Conv-Scale}_{M,2} := \{\mathbf{y} = \sum_{j=1}^{M}\gamma_{j}\mathbf{W}_{j}\mathbf{x}|\mathbf{W}_{j} \in \mathbb{R}^{o,i}, \gamma_{j} \in \mathbb{R}^{o}\}.
\end{equation}
The end-to-end weight $\mathbf{W}_{e_2,cs} := \sum_{j=1}^{M}\gamma_{j}\mathbf{W}_{j}$ is updated by:
\begin{equation}
\label{eq:nlsp}
\begin{split}
\mathbf{W}^{(t+1)}_{e_2,cs} :=  \mathbf{W}^{(t)}_{e_2,cs} - &\eta \sum_{j=1}^{M} (vec(diag(\mathbf{W}_j^{(t)})^{2}) +\left \| \gamma_j^{(t)} \right \|_2^2)\mathbf{x}^{\top}\frac{\partial L}{\partial \mathbf{y}} + O(\eta^2).
\end{split}
\end{equation}
With the same precondition $\mathbf{W}^{(t)}_{cs} = \mathbf{W}^{(t)}_{e_2,cs}$, Eqn. \eqref{eq:nlsp} will never be equivalent to Eqn. \eqref{eq:sgdcn} when Condition \ref{cond:act} is satisfied:\\

\begin{condition} \label{cond:act}
	At least two of all the branches are active.
	\begin{equation}
	\begin{split}
	\exists \;& \mathbf{S} \subseteq \{1, 2, \cdots, M\}, \left | \mathbf{S}  \right | \ge 2, such \; that \; \forall j \in \mathbf{S} ,\;   vec(diag(\mathbf{W}_j^{(t)})^{2}) +\left \| \gamma_j^{(t)} \right \|_2^2 \ne \mathbf{0}.
	\end{split}
	\end{equation}
\end{condition}
To imply the conclusion above, first we notice the square form of precondition that:
\begin{equation}
\label{eq:2foreq}
\gamma \mathbf{W} = \sum_{j=1}^{M}\gamma_{j}\mathbf{W}_{j} \implies \gamma^2 vec(diag(\mathbf{W})^{2}) = (\sum_{j=1}^M \gamma_{j} vec(diag(\mathbf{W}_j)))^2.
\end{equation}
Meanwhile, if Eqn. \eqref{eq:nlsp} and \eqref{eq:sgdcn} were equivalent, we have:
\begin{equation}
\label{eq:2backeq}
\gamma^2 vec(diag(\mathbf{W})^{2}) = \sum_{j=1}^M \gamma_{j}^2 vec(diag(\mathbf{W}_j)^2).
\end{equation}
By subtracting Eqn. \eqref{eq:2foreq} and \eqref{eq:2backeq}, we come to the following result:
\begin{equation}
\label{eq:2sub}
\sum_{i=1}^M\sum_{j=1,j\ne i}^M \gamma_{i}\gamma_{j} vec(diag(\mathbf{W}_i)diag(\mathbf{W}_j)) = \mathbf{0}.
\end{equation}
This indicates either (1) there is strictly no correlation across all branches, which is not practical. (2) or at most one branch is active, which contradicts Condition \ref{cond:act}.\\
\begin{condition} \label{cond:ne}
	The initial state of each active branch is different from that of each other.
	\begin{equation}
	\forall j_1, j_2 \in \mathbf{S}, \; j_1 \ne j_2, \quad	\mathbf{W}_{j_1}^{(0)} \ne \mathbf{W}_{j_2}^{(0)}.
	\end{equation}	
\end{condition}
Meanwhile, when Condition \ref{cond:ne} is met, the multi-branch structure will not degrade into single one for both forwarding:
\begin{equation}
\gamma_{j_1}\mathbf{W}_{j_1} \ne \gamma_{j_2}\mathbf{W}_{j_2} \iff \gamma_{j_1}\mathbf{W}_{j_1}\mathbf{x} \ne \gamma_{j_2}\mathbf{W}_{j_2}\mathbf{x}
\end{equation}
and backwarding:
\begin{equation}
vec(diag(\mathbf{W}_{j_1})^{2}) +\left \| \gamma_{j_1} \right \|_2^2 \ne vec(diag(\mathbf{W}_{j_2})^{2}) +\left \| \gamma_{j_2} \right \|_2^2 \iff \frac{\partial L}{\partial (\gamma_{j_1}\mathbf{W}_{j_1})} \ne \frac{\partial L}{\partial (\gamma_{j_2}\mathbf{W}_{j_2})},
\end{equation}
which reveals the following proposition explaining why the scaling factors are important.  Note that both Condition \ref{cond:act} and \ref{cond:ne} are always met when weights $\mathbf{W}_{j}^{(0)}$ of each branch is random initialized \cite{he15} and scaling factors $\gamma_{j}^{(0)}$ are initialized to 1. 

\begin{proposition}
	\label{prop}
	A single-branch linear mapping, when re-parameterizing parts or all of it by over-two-layer multi-branch topologies, the entire end-to-end weight matrix will be differently optimized. If one layer of the mapping is re-parameterized to up-to-one-layer multi-branch topologies, the optimization will remain unchanged.
\end{proposition} 



So far, we have extended the discussion on how re-parameterization impacts optimization, from multi-layer only \cite{Arora18} to multi-branch included as well. Actually, all current effective re-parameterization topology \cite{Ding19, Ding21dbb, Ding21repvgg, Guo20, Cao20} can be validated by either Lemma \ref{lemma} or Proposition \ref{prop}.

\begin{figure*}[bp]
	\centering
	\includegraphics[width=0.9\textwidth]{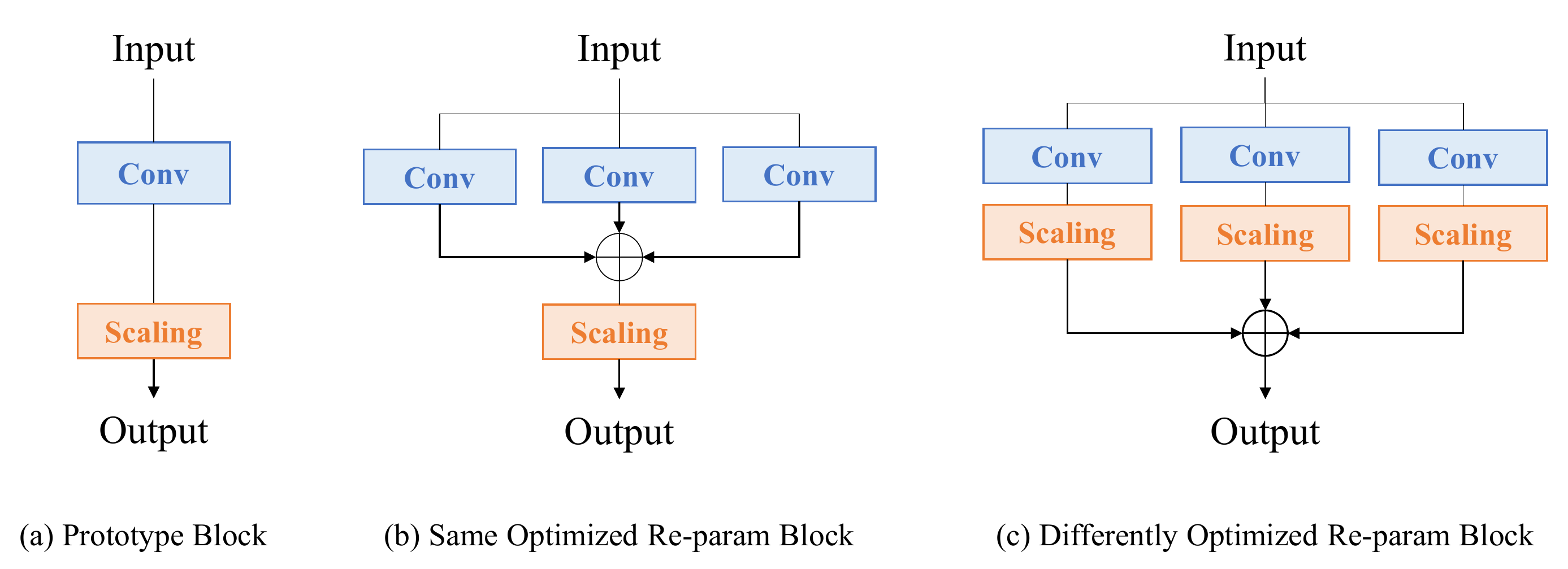}
	\caption{An example for illustrating Proposition \ref{prop}.}
	\label{fig:grad}
\end{figure*}

\paragraph{The impacts of momentum and weight decay.} Momentum and weight decay are often deployed in company with SGD.  updating of a local multi-branch multi-layer topology $\Phi$ with $M$ branches and $N_m$ in each branch $m$.

\begin{equation}
\begin{split}
\Phi := \{\mathbf{y} = \sum_{m=1}^{M}\prod_{n=1}^{N_m}\mathbf{W}_{mn} \mathbf{x} \triangleq \sum_{m=1}^{M}  \mathbf{W}_{m}\mathbf{x}  |\mathbf{W}_{mn} \in \mathbb{R}^{d_{mn}\times d_{m(n-1)}}\}.
\end{split}
\end{equation}

Each weight instance $\mathbf{W}_{mn}$ is optimized by SGD with a learning rate $\eta$, a weight decay $\lambda$, and a momentum coefficient $\mu$:

\begin{equation}
\mathbf{W}_{mn}^{(t+1)} := (1-\eta\lambda)\mathbf{W}_{mn}^{(t)} - \eta\sum_{\tau=1}^{t}(\eta\mu)^{t-\tau}\frac{\partial L}{\partial \mathbf{W}_{mn}^{(\tau)}}, 
\end{equation}

leading to an update of the end-to-end mapping $\mathbf{W}_{e}$ of:

\begin{equation}
	\label{eq:update}
	\begin{split}
	\mathbf{W}_{e}^{(t+1)} = \sum_{m=1}^{M}(1-\eta\lambda N_m)  \mathbf{W}_{m}^{(t)} -  \eta \sum_{\tau=1}^{t}(\eta\mu)^{t-\tau}\sum_{m=1}^{M} \left \| \mathbf{W}_m^{(t)} \right \|_2^{2-\frac{2}{N_m}}   	\cdot  ({(\mathbf{x}^{\top}\frac{\partial L}{\partial \mathbf{y}})}^{(t)} + (N_m-1) \cdot Pr_{\mathbf{W}_m^{(t)}}{\{\mathbf{x}^{\top}\frac{\partial L}{\partial \mathbf{y}}\}}^{(t)}) + O(\eta^2),
	\end{split}
\end{equation}
where $Pr_{\mathbf{W}}\{\mathbf{G}\} $ is the projection of $\mathbf{G}$ the direction of $\mathbf{W}$ defined in Eqn. \eqref{eq:proj}. As the condition in Proposition \ref{prop} goes, if one layer is re-paramed to multi-branches with a maximum depth of one, the weight decay item will remain unchanged if $N_m=1$ for all $m$. However, a minor difference may exist, considering that the introduction of zero-depth (constant) items means different $\mathbf{W_m}^{(t)}$ with a same $\mathbf{W_e}^{(t)}$, which further changes the decayed part of the $t$ moment weight. This indicates optimization differences may occur in identically initialized layers with a vanilla or a re-parameterized implementation. Meanwhile, the momentum items can be regarded as previous gradients, thus having no impact on the correctness of Proposition \ref{prop}. 

%% file: sections_supp/experiment.tex
\section{Experimental Results}

\subsection{Implementation Details}
\paragraph{Initialization of branch-wise scaling layers.} During the linearization step in Sec. 3.2 of the body, the original branch-wise norm layers replaced with scaling layers. Therefore, the branch-wise numerical stability provided by norm layers is no longer maintained. To balance the distribution of output feature maps of each branch, we have to carefully initialize the weights of scaling layers. Specifically, the scaling factors are initialized to \{1.0, 0.25, 0.5, 0.5, 0.0, 0.5\} for the \{1$\times$1, k$\times$k, 1$\times$1-k$\times $k, 1$\times$1-pooling, 1$\times$1-filtering, dw-pw conv\} branches respectively. There are several intuitive reasons for such an initialization strategy: (1) The scaling factors after 1$\times$1 conv is set the highest among scaling layers of all branches. This is because the variance of feature maps  convoluted by 1$\times$1 sized kernels is generally much smaller then those convoluted by k$\times$k sized kernels. (2) We set larger scaling factors for the 1$\times$1-k$\times $k branch than the k$\times $k branch (0.5$>$0.25), since both branches are similar spatial-channel correlational layers, and that the 1$\times$1-k$\times $k branch contains more parameters. (3) We decrease the scaling factors of the 1$\times$1-pooling and 1$\times$1-filtering branches, as weights of these branches are easily to be trapped in shallow local optima during early stages of training. This will in turn suppress the representative of other branches, which are harder to converge. Here we only make a very limited number of attempts to initializing with different values. Meanwhile, neither brute force searching methods nor differential ones \cite{Zhang21} are deployed.

\paragraph{Data pre-processing.} Our experimental results for ResNets and DBB-ResNets on ImageNet is different from that of the literature of DBB \cite{Ding21dbb}. This is mainly owned to a different data pre-processing pipeline. Actually, we process the input batches exactly the same as Ding \etal in RepVGG \cite{Ding21repvgg}. Compared to the DBB pre-processing setting, the performance improvement brought by re-parameteriaztion methods becomes marginal. However, all models achieve higher accuracy and that's why we change the setting.


\subsection{More Results}
\paragraph{Re-param on resnet variants.} Depth and width are two different dimensions of model capacity \cite{Tan19,Kornblith21}. As larger ResNet models only extends in the depth, we further conduct experiments on wider ResNets \cite{Zagoruyko17b}. From Table \ref{tab:wresnet} we show that OREPA generalize well on wider neural networks. Besides, ResNeXts\cite{Xie17} share very similar architectures with bottleneck-ResNets, thus benefiting from OREPA in the same way as ResNets do, as presented in Table \ref{tab:wresnet}.

\begin{table}[htbp]
	\centering
	\caption{Experiments on ImageNet for 120 epochs for ResNet variants. All models reported are trained with batch size 256 on our machine with 4 Nvidia Tesla V100 (32G) GPUs for a fairer comparison.}
	\vspace{0.5em}
	\begin{tabular}{@{}cccc|ccc|ccc@{}}
		\toprule
		\multirow{3}{*}{Re-param} & \multicolumn{3}{c|}{ResNeXt-50} & \multicolumn{3}{c|}{WideResNet-18 (1.5$\times$)} & \multicolumn{3}{c}{WideResNet-18 (2$\times$)} \\ \cmidrule(l){2-10} 
	    & Top1- & GPU- & Training &  Top1- & GPU- & Training & Top1- & GPU- & Training \\
		& Acc & Mem & time/batch & Acc & Mem & time/batch & Acc & Mem & time/batch  \\ \midrule
		None & 77.14 & 12.4G & 0.286s & 73.69 & 6.0G & 0.121s & 74.74 & 7.4G & 0.161s \\
		OREPA & 77.66 & 13.0G & 0.343s & 74.51 & 7.4G & 0.164s & 75.61 & 10.4G & 0.229s
	\end{tabular}
	\label{tab:wresnet}
\end{table}

\begin{figure*}[htbp]
	\centering
	\includegraphics[width=0.95\textwidth]{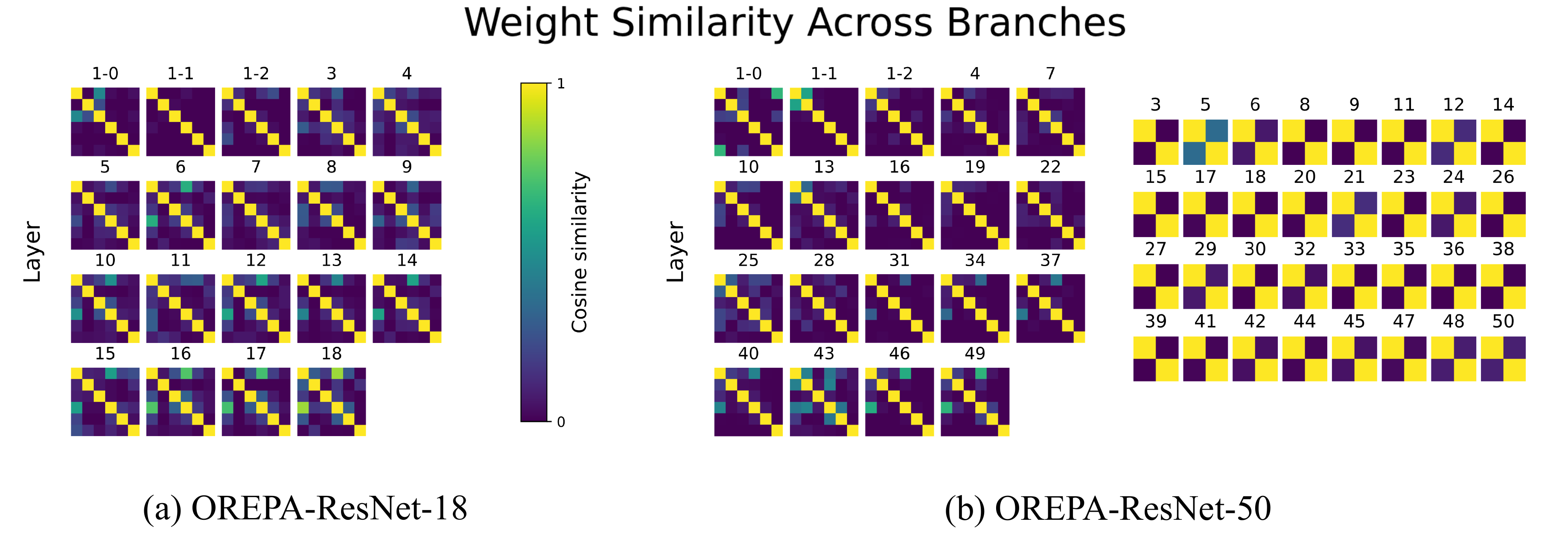}
	\caption{Visualization of branch-level similarity. We calculate
		cosine similarities between the weights from different branches. The mid and right plots respectively correspond to $3\times3$ and $1\times1$ convolutional layers in ResNet-50.}
	\label{fig:viz}
\end{figure*}

\begin{figure*}[htbp]
	\centering
	\includegraphics[width=1.0\textwidth]{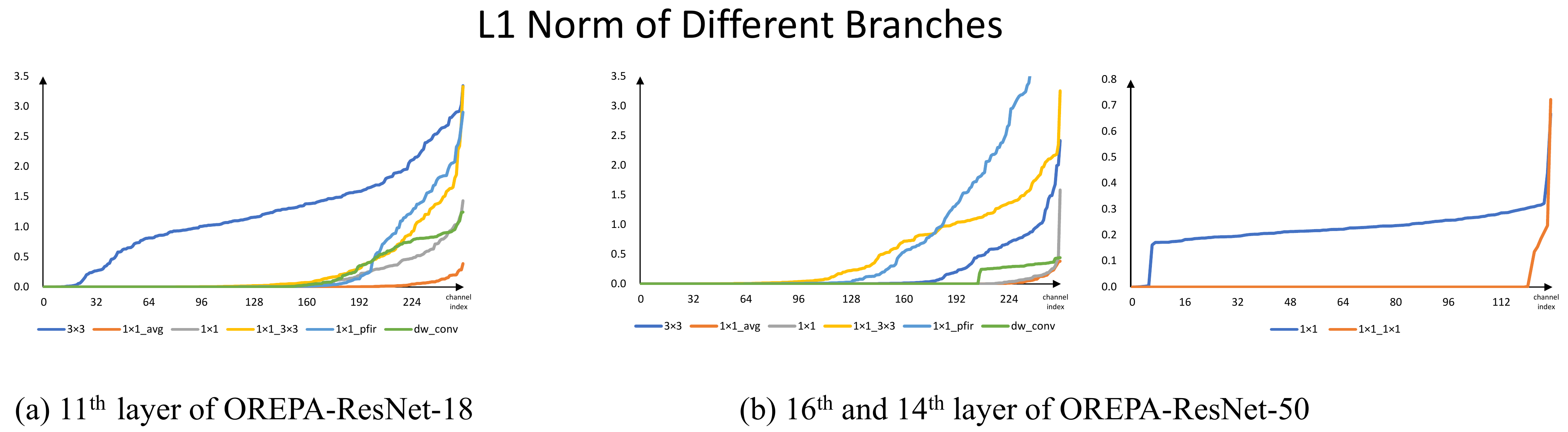}
	\caption{Sorted normalized norms of different branches. We notice that the average contribution of the branches varies. However, each branch matters for some specific channels.}
	\label{fig:viz_norm}
\end{figure*}

\paragraph{More visualization results.} In Figure \ref{fig:viz}, we visualize branch-wise similarity of all the branches in more models and blocks. It is clear that all branches are diversely optimized. We further presented the norm of branch-wise kernels for each output channel in Figure \ref{fig:viz_norm}. The branches do not contribute to the squeezed kernel equally. However, each of them is important for some specific channels.

\subsection{Object Detection and Semantic Segmentation: Detailed Setup}
In this part, we describe the detailed experimental setup for object detection and semantic segmentation.
For these two tasks, we apply the commonly used models~\cite{Ren15, Lin17, Zhao17, Chen18} 
and only replace the backbone with the re-param ones pretrained on ImageNet.

\paragraph{Object Detection}
We conduct experiments on the popular MS-COCO~\cite{Lin14} dataset to test the performances on object detection.
Following the common practice, we use the COCO \textit{trainval35k} split~(115K images) for training 
and \textit{minival}~(5K images) for validation.
We use both the two-stage model, Faster R-CNN~\cite{Ren15}, and the one-stage model, RetinaNet~\cite{Lin17} 
with ResNet-50~(DBB-50, OREPA-ResNet-50) as the backbones. 
By default, we apply an SGD optimizer with 0.02 as the initial learning rate.
The batch size is set to 4. We apply the 1$\times$ schedule, where the 
total training epoch is 12 and the learning rate is reduced by 10 
after the $8^{th}$ and $11^{th}$ epochs. 
The shorter sides of images is set to 800 and the longer ones are less than 1333.
For our OREPA-ResNet, we do not freeze the first stage~(linear deep stem).
Instead, we reduce the corresponding learning rate by 10.
All of our experiments are conducted using mmdetection~\cite{mmdetection}.

\paragraph{Semantic Segmentation}
For semantic segmentation, we choose PSPNet~\cite{Zhao17} and DeepLabV3+~\cite{Chen18} 
with ResNet-50~(DBB-50, OREPA-ResNet-50) as the backbones.
We evaluate the models on Cityscapes~\cite{Cordts16}.
It contains 5000 high-resolution finely annotated images, which are divided into 
2975, 500, and 1525 for training, validation and testing. 
There are 19 classes in total for training and evaluation.
We use the SGD optimizer with initial learning rate 0.02 to 
train for 40k iterations. The poly learning rate with power 0.9 and 
minimum learning rate 2e-4 is applied.
The default batch size is set to 2 and the weight decay is 5e-4.
Training data augmentation includes random scaling between 0.5$\times$ to 2.0$\times$, 
random horizontal flipping, random cropping to [512, 1024].
By default, we DO NOT apply the Sync-BatchNorm for all the backbones.
In the inference phase, we report the accuracy~(mIoU) on the validation set
without any test-time augmentation. All of our experiments are conducted using mmsegmentation~\cite{mmseg2020}.

\subsection{Limitations: Discussion on Residual awareness.} The residual connection \cite{He16} is believed important for training very deep neural networks for alleviate the gradient vanishing problem \cite{Balduzzi17}. Recently, RepVGG \cite{Ding21repvgg} is proposed as a high performance residual-free architecture. Inside each building block of RepVGG, a training-time identity branch is maintained for provide residual connections. During inference, such an identity branch can be squeezed into a convolutional layer for faster inference. Can such identity branches be online re-parameterized into one layer at the training stage? We find it hard to give an affirmative answer. On the one hand, the post-identity-addition norm layer has been proved an inferior design experimentally \cite{He16b}. On the other hand, we conjecture that the training-time \textbf{\emph{branch-wise non-linearity}} brought by norm layers is \textbf{\emph{more important in residual-free}} (VGG-like) architectures then residual-based ones. Based on this point, we reserve all three branches in RepVGG blocks instead of squeezing them into one.

\begin{figure*}[htbp]
	\centering
	\subfigure{\includegraphics[width=0.45\textwidth]{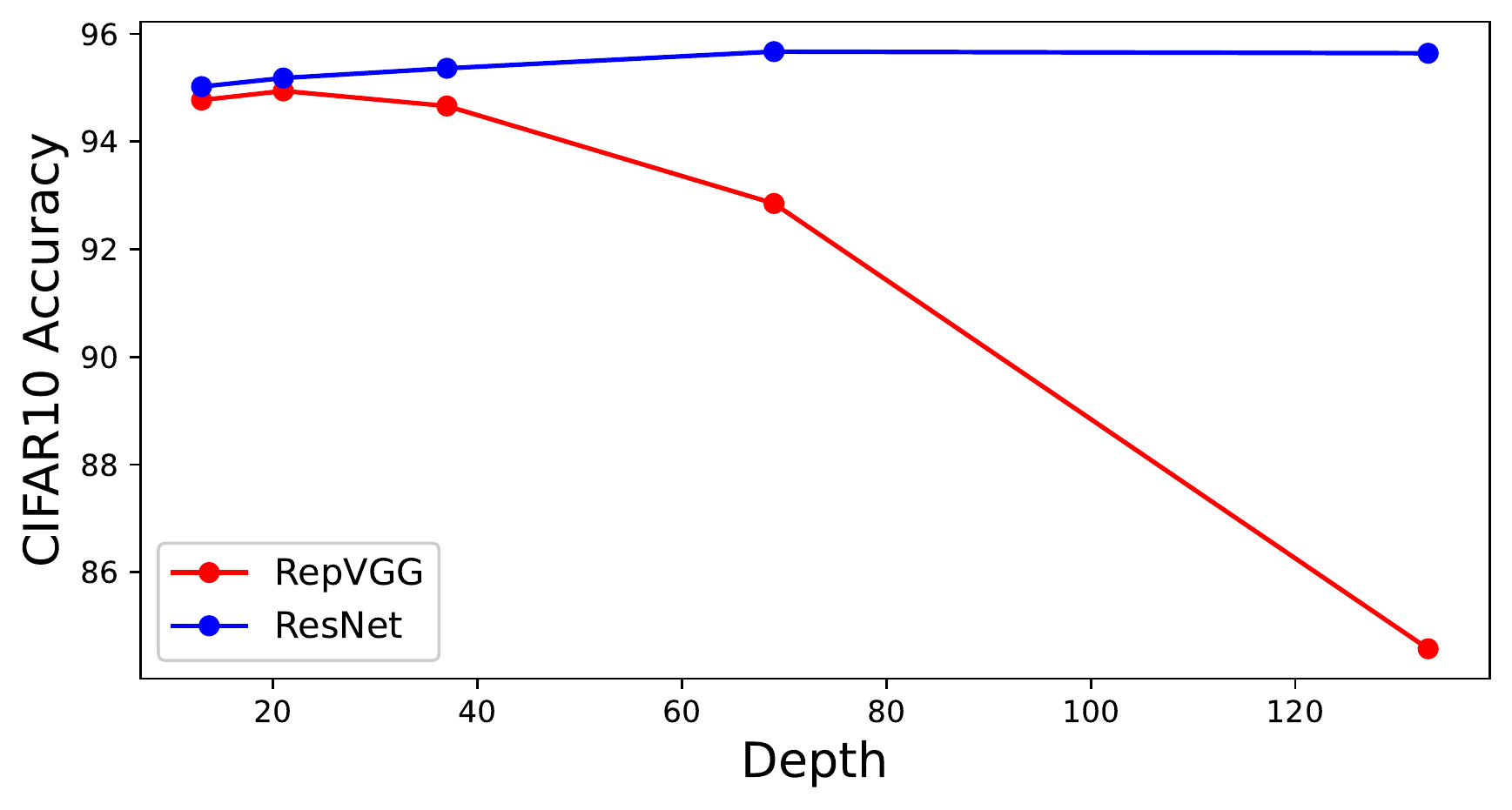}}
	\subfigure{\includegraphics[width=0.45\textwidth]{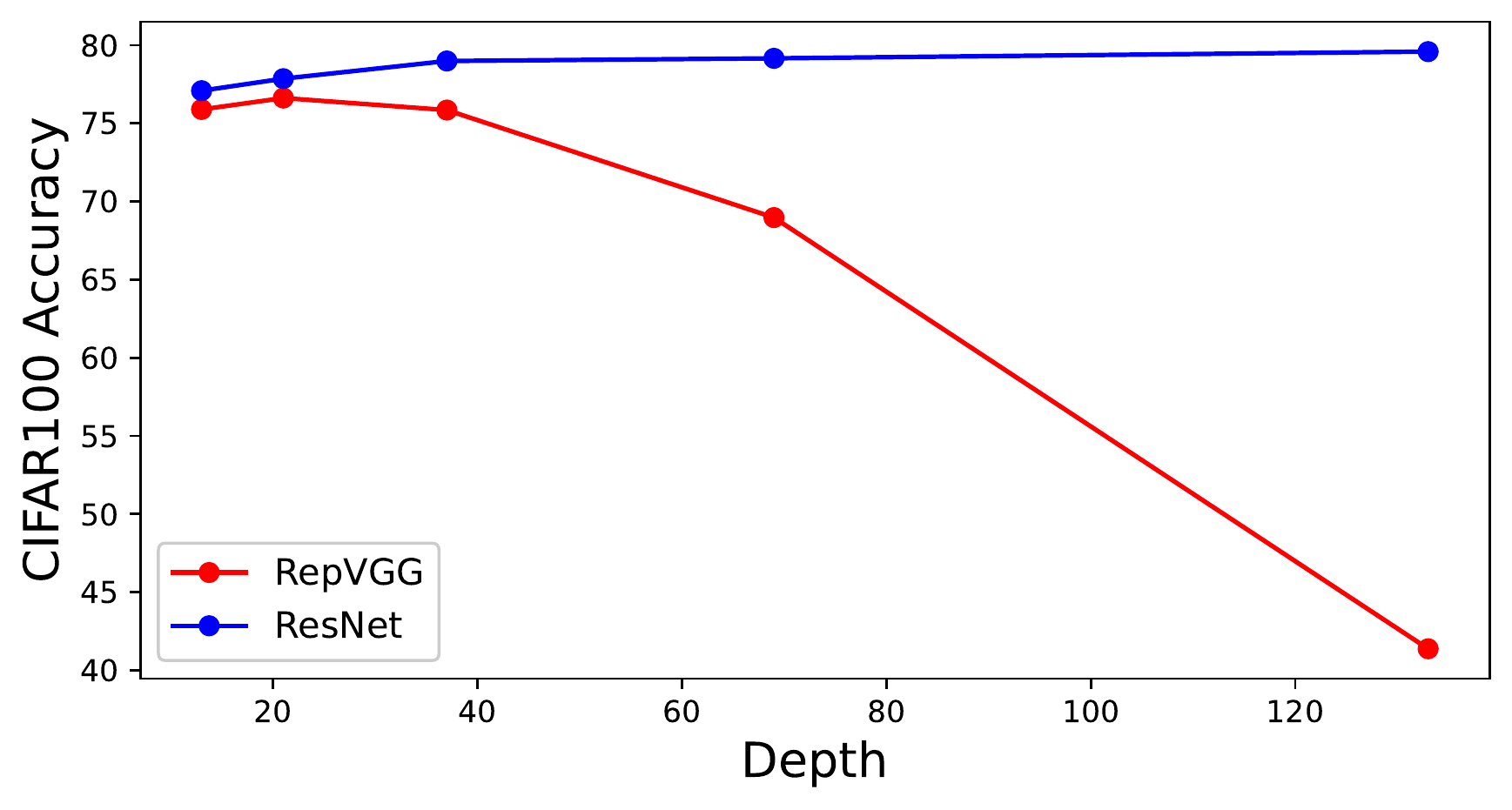}}
	\caption{\textbf{Figure directly copied from \cite{Meng21}}. When going deeper, RepVGG models suffer performance degradation while ResNets do not.}
	\label{fig:deeper-repvgg-and-resnet}
\end{figure*}

To understand the inconsistent significance of norm layers for architectures with different residual-awareness, let us come to Fig 
\ref{fig:deeper-repvgg-and-resnet} \cite{Meng21} first. It is obvious that RepVGG models have difficulty in going deeper, unlike ResNets. This indicates that the quasi-residual connections provided by identity branches in RepVGG is not the same as regular residual ones. More specifically, the identity branches in RepVGG do not connect feature maps across activations, while those in ResNet do. Based on these observations, we assume that, in residual-based architectures the training-time branch-wise non-linearity in re-parameterization blocks are less important, as such properties have been provided by residual connections. However, it is not the case in residual-free architectures.

There are three branches of identity, $1\times1$ conv, and $3\times3$ conv in the RepVGG blocks, as presented in Table \ref{tab:ra0res} below. We \textbf{\emph{do not online}} merge the three branches for mainly two reasons: (1) There is little space for \textbf{\emph{resource}} optimization in RepVGG blocks (None vs. RepVGG), since identity and $1\times1$ conv are light weighted. (2) As has been stated in the Supp. Sec. 2.4, the branches in RepVGG blocks are critical for providing approximate \textbf{\emph{residual}} connections (RepVGG-Online vs. RepVGG). This could be related to intrinsic discrepancies of residual-based/free architectures, which is still a important topic for the community.

\begin{figure*}[htbp]
	\centering
	\includegraphics[width=1.0\textwidth]{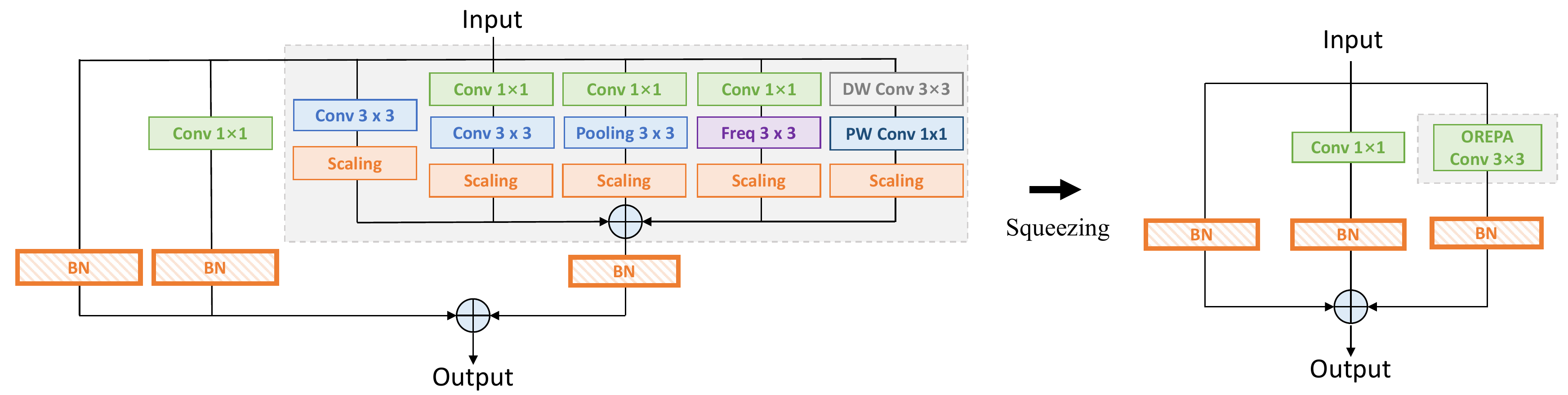}
	\caption{The design of the proposed OREPA-VGG block during training.}
\end{figure*}

Actually, results reported in Table 4 include additional branches online merged into the $3\times3$ conv. Compared to an offline counterpart (OREPAVGG-Offline), online re-param saves extra GPU memory by 97\% and accelerates training for 4$\times$, as shown in Tab. \ref{tab:ra0res} While RepVGG boosts residual awareness in VGGs, can we explore beyond with more complex structures? Our work shed light on building complicated topologies with as little addition costs as possible for re-param community, despite OREPAVGG might not be the optimal structure.

Even though, we still don't need to worry about the generalization of the proposed OREPA, considering that most architectures are residual-aware \cite{Xie17, Tan19, Brock21nfnet, Alexey21, Radosavovic20}. Even for residual-free architectures like RepVGG, OREPA provides solutions for trading-off between model augmentation and extra training costs. More importantly, online re-parameterization make it possible to build very deep and wide training-time blocks. We leave the systematical exploration towards more effective re-parameterization for future works and other researchers. 

\begin{table*}[hbp]
		\caption{Results on all re-param variants for the model RepVGG-A0. In the ``Structure" column, $\surd$ indicate explicit layers with a norm layer following, while $\in 3\times3$ represent implicit ones re-paramed into the $3\times3$ layer.}
		\resizebox{1.00\textwidth}{!}{
			\begin{tabular}{@{}cccccccccccc@{}}
				\toprule
				\multirow{2}{*}{Model} & \multirow{2}{*}{Re-param} & \multicolumn{7}{c}{Structure} & Top1- & GPU- & Training \\  
				& & 3$\times$3 & identity & 1$\times$1 & 1$\times$1-3$\times$3 & 1$\times$1-avg & 1$\times$1-freq\_fir & dw3$\times$3-1$\times$1($8\times$) & Acc & Mem & time/batch \\ \midrule
				
				\multirow{5}{*}{RepVGG-A0}
				& \textbf{None} & $\surd$ &  &  & &  &  &  & 71.17 & 3.4G & 0.083s  \\
				& RepVGG-Online & $\surd$ & $\ast$ & $\ast$  & &  &  &  & 71.90 & 3.4G & 0.086s  \\
				& RepVGG & $\surd$ & $\surd$ & $\surd$ &  &  & &  & 72.41 & 3.8G & 0.100s \\
				
				& OREPAVGG (reported) & $\surd$ & $\surd$ & $\surd$ &  $\ast$ &  $\ast$ & $\ast$ & $\ast$ &  73.04 & 4.2G & 0.136s \\
				& OREPAVGG-Offline & $\surd$ & $\surd$ & $\surd$ & $\surd$ & $\surd$ & $\surd$ & $\surd$ & 72.96 & 16.1G & 0.538s \\
		\end{tabular}}
		\label{tab:ra0res}
\end{table*}